\documentclass{article}

\PassOptionsToPackage{numbers,sort&compress}{natbib}

\usepackage[final]{neurips_2023}

\usepackage[utf8]{inputenc} %
\usepackage[T1]{fontenc}    %
\usepackage{booktabs}       %
\usepackage{amsfonts}       %
\usepackage{nicefrac}       %
\usepackage{xfrac}
\usepackage{microtype}      %
\usepackage{bm} %
\usepackage{amsmath}
\usepackage[dvipsnames]{xcolor}
\usepackage{numprint}
\npdecimalsign{.}
\usepackage{caption}
\usepackage{subcaption}
\usepackage[export]{adjustbox}
\usepackage{makecell}
\usepackage{multirow}
\usepackage{xspace}
\usepackage{amssymb}
\usepackage{enumitem}
\usepackage{mathtools}
\usepackage{listings}
\let\originalleft\left
\let\originalright\right
\renewcommand{\left}{\mathopen{}\mathclose\bgroup\originalleft}
\renewcommand{\right}{\aftergroup\egroup\originalright}

\newcommand{\Fig}[1]{\hyperref[{#1}]{Figure~\ref*{#1}}}  %
\newcommand{\fig}[1]{\hyperref[{#1}]{Fig.~\ref*{#1}}}    %
\newcommand{\Figs}[2]{\hyperref[{#1}]{Figures~\ref*{#1}} and \ref{#2}} 
\newcommand{\Tab}[1]{\hyperref[{#1}]{Table~\ref*{#1}}}
\newcommand{\tab}[1]{\hyperref[{#1}]{Table~\ref*{#1}}}
\newcommand{\Eqn}[1]{\hyperref[{#1}]{Equation~\ref*{#1}}}
\newcommand{\eqn}[1]{\hyperref[{#1}]{Eq.~\ref*{#1}}} %
\newcommand{\eqnp}[1]{(Eq.~\ref{#1})} %
\newcommand{\sect}[1]{\hyperref[{#1}]{Sec.~\ref*{#1}}} %
\newcommand{\supp}[1]{\hyperref[{#1}]{Suppl.~\ref*{#1}}}
\newcommand{\app}[1]{\hyperref[{#1}]{App.~\ref*{#1}}}
\newcommand{\App}[1]{\hyperref[{#1}]{Appendix~\ref*{#1}}}

\usepackage{xspace}
\makeatletter
\DeclareRobustCommand\onedot{\futurelet\@let@token\@onedot}
\def\@onedot{\ifx\@let@token.\else.\null\fi\xspace}
\def\eg{e.g\onedot}
\def\ie{i.e\onedot}

\def\vs{vs\onedot}
\makeatother

\usepackage{bm}
\def\vf{{\bm{f}}}
\def\vh{{\bm{h}}}
\def\vm{{\bm{m}}}
\def\vp{{\bm{p}}}
\def\vs{{\bm{s}}}

\def\vx{{\bm{x}}}
\def\vy{{\bm{y}}}

\def\mA{{\bm{A}}}
\def\mP{{\bm{P}}}

\usepackage{xr-hyper}
\makeatletter
\newcommand*{\addFileDependency}[1]{%
  \typeout{(#1)}
  \@addtofilelist{#1}
  \IfFileExists{#1}{}{\typeout{No file #1.}}
}
\makeatother

\newcommand{\method}{VideoSAUR\xspace}

\DeclareMathOperator{\SA}{SA}
\DeclareMathOperator{\CE}{CE}

\newcommand{\paramsdec}{\psi}
\usepackage{wrapfig}
\usepackage{titletoc}
\usepackage{hyperref}
\usepackage{url}
\hypersetup{colorlinks,
            linkcolor=NavyBlue,
            urlcolor=NavyBlue,
            citecolor=NavyBlue}

\definecolor{codegreen}{rgb}{0,0.6,0}
\definecolor{codegray}{rgb}{0.5,0.5,0.5}
\definecolor{codepurple}{rgb}{0.58,0,0.82}
\definecolor{backcolour}{rgb}{0.95,0.95,0.92}
\lstdefinestyle{mystyle}{
    backgroundcolor=\color{backcolour},   
    commentstyle=\color{codegreen},
    keywordstyle=\color{magenta},
    numberstyle=\tiny\color{codegray},
    stringstyle=\color{codepurple},
    basicstyle=\ttfamily\scriptsize,
    breakatwhitespace=false,         
    breaklines=true,                 
    captionpos=b,                    
    keepspaces=true,                 
    numbers=left,                    
    numbersep=5pt,
    showspaces=false,                
    showstringspaces=false,
    showtabs=false,                  
    tabsize=2
}
\title{
Object-Centric Learning for Real-World Videos by Predicting Temporal Feature Similarities
}

\author{Andrii Zadaianchuk$^{1,2}$\thanks{equal contribution}\quad\quad\quad Maximilian Seitzer$^{1 *}$\quad\quad\quad Georg Martius$^{1}$ \\
$^1$ Max Planck Institute for Intelligent Systems, T\"ubingen, Germany\\
$^2$ Department of Computer Science, ETH Zurich\\
\texttt{andrii.zadaianchuk@tuebingen.mpg.de} \\
}

\begin{document}

\maketitle

\begin{abstract}
Unsupervised video-based object-centric learning is a promising avenue to learn structured representations from large, unlabeled video collections, but previous approaches have only managed to scale to real-world datasets in restricted domains.
Recently, it was shown that the reconstruction of pre-trained self-supervised features leads to object-centric representations on unconstrained real-world image datasets.
Building on this approach, we propose a novel way to use such pre-trained features in the form of a temporal feature similarity loss.
This loss encodes semantic and temporal correlations between image patches and is a natural way to introduce a motion bias for object discovery.
We demonstrate that this loss leads to state-of-the-art performance on the challenging synthetic MOVi datasets.
When used in combination with the feature reconstruction loss, our model is the first object-centric video model that scales to unconstrained video datasets such as YouTube-VIS.
\texttt{\href{https://martius-lab.github.io/videosaur/}{https://martius-lab.github.io/videosaur/}}
\end{abstract}

\section{Introduction}

Autonomous systems should have the ability to understand the natural world in terms of independent entities. %
Towards this goal, unsupervised object-centric learning methods~\citep{burgess2019monet,greff2019multi,Locatello2020SlotAttention} learn to structure scenes into object representations solely from raw perceptual data.
By leveraging large-scale datasets, these methods have the potential to obtain a robust object-based understanding of the natural world.
Of particular interest in recent years have been video-based methods~\cite{Jiang2020SCALOR,kipf2022conditional,elsayed2022savi++,Singh2022STEVE}, not least because the temporal information in video presents a useful bias for object discovery~\cite{Bao2022ObjectsThatMove}.
However, these approaches are so far restricted to data of limited complexity, successfully discovering objects from natural videos only on closed-world datasets in restricted domains.

In this paper, we present the method {\it{\bf\emph{Video}} {\bf\emph S}lot {\bf\emph A}ttention {\bf\emph U}sing temporal feature simila{\bf\emph R}ity}, \method, that scales video object-centric learning to unconstrained real-world datasets covering diverse domains.
To achieve this, we build upon recent advances in image-based object-centric learning.
In particular, \citet{seitzer2023bridging} showed that reconstructing pre-trained features obtained from self-supervised methods like DINO~\cite{Caron2021DINO} or MAE~\cite{He2022MAE} leads to state-of-the-art object discovery on complex real-world images.
We demonstrate that combining this feature reconstruction objective with a video object-centric model~\cite{kipf2022conditional} also leads to promising results on real-world YouTube videos.

\begin{figure}
    \centering
    \includegraphics[width=\textwidth]{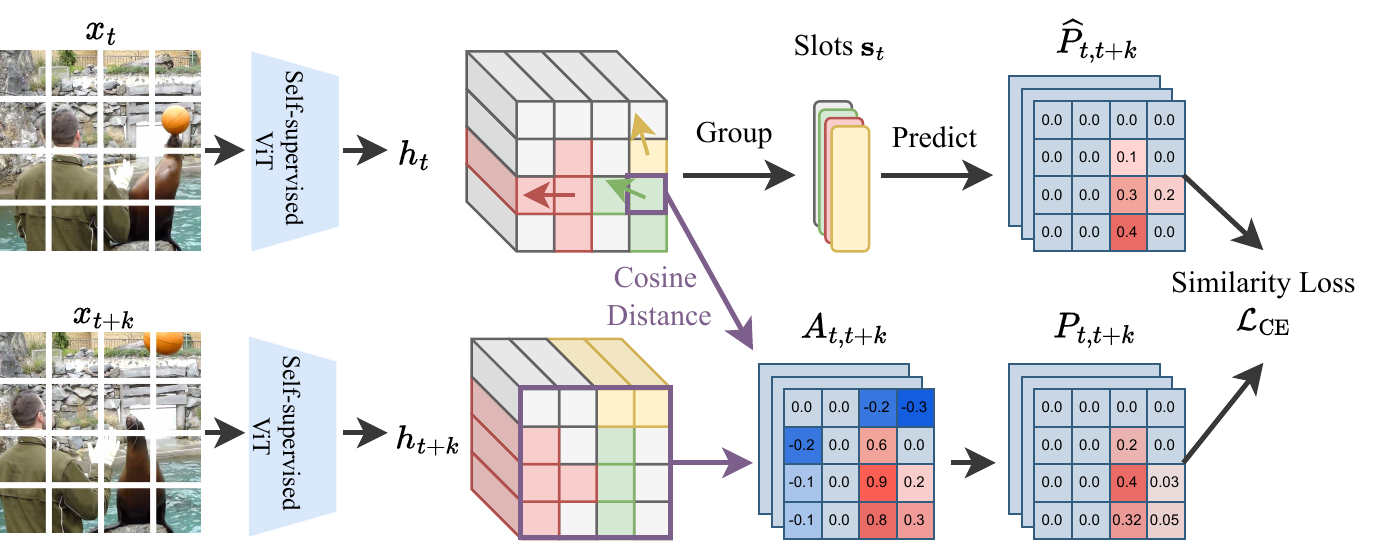}
    \caption{We propose a \emph{self-supervised temporal similarity loss} for training object-centric video models. 
    For each patch at time $t$, the model has to predict a distribution $\hat{\mP}_{t, t+k}$ indicating where all semantically-similar patches have moved to $k$ steps into the future.
    The target distribution $\mP_{t, t+k}$ is computed with a softmax on the affinity matrix $\mA_{t, t+k}$ containing the cosine distance between all patch features $\vh_t$, $\vh_{t+k}$.
    The loss incentivizes the model to group areas with consistent motion and semantics into slots.}
    \label{fig:feature-sim-loss}
\end{figure}

We then identify a weakness in the training objective of current unsupervised video object-centric architectures~\cite{Jiang2020SCALOR,Singh2022STEVE}: the prevalent reconstruction loss does not exploit the temporal correlations existing in video data for object grouping.
To address this issue, we propose a novel self-supervised loss based on \emph{feature similarities} that explicitly incorporates temporal information (see~\fig{fig:feature-sim-loss}).
The loss works by predicting distributions over similarities between features of the current and future frames.
These distributions encode information about the motion of individual image patches.
To efficiently predict those motions through the slot bottleneck, the model is incentivized to group patches with similar motion into the same slot, leading to better object groupings as patches belonging to an object tend to move consistently.
In our experiments, we find that such a temporal similarity loss leads to state-of-the-art performance on challenging synthetic video datasets~\cite{Greff2021Kubric}, and significantly boosts performance on real-world videos when used in conjunction with the feature reconstruction loss.\looseness-1

In video processing, model efficiency is of particular importance. 
Thus, we design an efficient object-centric video architecture by adapting the SlotMixer decoder~\cite{sajjadi2022object} recently proposed for 3D object modeling for video decoding.
Compared to previous decoder designs~\cite{Locatello2020SlotAttention}, the SlotMixer decoder scales gracefully with the number of slots, but has a weaker inductive bias for object grouping.
We show that this weaker bias manifests in optimization difficulties in conjunction with conventional reconstruction losses, but trains robustly with our proposed temporal similarity loss.
We attribute this to the \emph{self-supervised nature} of the similarity loss:  compared to reconstruction, it requires predicting information that is not directly contained in the input; the harder task seems to compensate for the weaker bias of the SlotMixer decoder.

To summarize, our contributions are as follows:
(1) we propose a novel self-supervised loss for object-centric learning based on temporal feature similarities,
(2) we combine this loss with an efficient video architecture based on the SlotMixer decoder where it synergistically reduces optimization difficulties,
(3) we show that our model improves the state-of-the-art on the synthetic MOVi datasets by a large margin, and
(4) we demonstrate that our model is able to learn video object-centric representations on the YouTube-VIS dataset~\cite{vis2021}, while staying fully unsupervised.
This paper takes a large step towards unconstrained real-world object-centric learning on videos.

\section{Related Work}

\textbf{Video Object-Centric Learning\quad}
There is a rich body of work on discovering objects from video, with two broad categories of approaches: tracking bounding boxes~\citep{Kosiorek2018SQAIR,crawford2020spair,Jiang2020SCALOR,Lin2020SPACE} or segmentation masks~\citep{greff2017neural,vanstennkiste2018relational,veerapaneni2019entity, greff2019multi,Zoran2021parts,weis2021benchmarking,kabra2021simone,kipf2022conditional,elsayed2022savi++,Singh2022STEVE,traub2023learning, safadoust2023multi}.
Architecturally, most recent image-based models for object-centric learning~\citep{Locatello2020SlotAttention,Singh2022SLATE,seitzer2023bridging} are based on an auto-encoder framework with a latent slot attention grouping module~\citep{Locatello2020SlotAttention} that extracts a set of slot representations.
For processing video data, a common approach~\citep{Zoran2021parts, kipf2022conditional,elsayed2022savi++,Singh2022STEVE,traub2023learning} is then to connect slots recurrently over input frames; the slots from the previous frame act as initialization for extracting the slots of the current frame.
We also make use of this basic framework.

\textbf{Scaling Object-Centric Learning\quad}
Most recent work has attempted to increase the complexity of datasets where objects can successfully be discovered, such as the synthetic ClevrTex~\citep{Karazija2021clevrtex} and MOVi datasets~\citep{Greff2021Kubric}.
On natural data, object discovery has so far been limited to restricted domains with a limited variety of objects, such as YouTube-Aquarium and -Cars~\citep{Singh2022STEVE}, or autonomous driving datasets like WaymoOpen or KITTI~\citep{Geiger2013KITTI}.
On more open-ended datasets, previous approaches have struggled~\citep{yang2022promising}.\looseness-1

To achieve scaling, some works attempt to \emph{improve the grouping module}, for example by introducing equivariances to slot pose transformations~\citep{Biza2023invariant}, smoothing attention maps~\citep{kim2023shepherding}, formulating grouping as graph cuts~\citep{pervez2023differentiable} or a stick-breaking process~\citep{engelcke2021genesis2}, or by overcoming optimization difficulties by introducing implicit differentiation~\citep{Chang2022implicitdiff,Jia2023boqsa}.
In contrast, we do not change the grouping module, but use the vanilla slot attention cell~\citep{Locatello2020SlotAttention}.

Another prominent approach is to introduce \emph{better training signals} than the default choice of image reconstruction.
For example, one line of work instead models the image as a distribution of discrete codes conditional on the slots, either autoregressively by a Transformer decoder~\citep{Singh2022SLATE,Singh2022STEVE}, or via diffusion~\citep{jiang2023object,wu2023slotdiffusion}.
While this strategy shows promising results on synthetic data, it so far has failed to scale to unconstrained real-world data~\citep{seitzer2023bridging}.

An alternative is to step away from fully-unsupervised representation learning by introducing \emph{weak supervision}.
For instance, SAVi~\citep{kipf2022conditional} predicts optical flow, and SAVi++~\citep{elsayed2022savi++} additionally predicts depth maps as a signal for object grouping.
Other works add an auxiliary loss that regularizes slot attention's masks towards the masks of moving objects~\citep{Bao2022ObjectsThatMove,bao2023motionguided}.
Our model also has a loss that focuses on motion information, but uses an unsupervised formulation.
OSRT~\citep{sajjadi2022object} shows promising results on synthetic 3D datasets, but is restricted by the availability of posed multi-camera imagery.
While all those approaches improve on the level of data complexity, it has not been demonstrated that they can scale to unconstrained real-world data.

The most promising avenue so far in terms of scaling to the real-world is to \emph{reconstruct features from modern self-supervised pre-training methods}~\citep{Caron2021DINO,He2022MAE,Assran2022MSN,Chen2021MoCov3}.
Using this approach, DINOSAUR~\citep{seitzer2023bridging} showed that by optimizing in this highly semantic space, it is possible to discover objects on complex real-world image datasets like COCO or PASCAL VOC.
In this work, we similarly use such self-supervised features, but for learning on video instead of images.
Moreover, we improve upon reconstruction of features by introducing a novel loss based on similarities between features.

\textbf{Concurrent Work\quad}
Parallel to this work, two more slot attention-based methods were proposed that learn object-centric representations on real-world videos: SMTC~\cite{qian2023semantics} and SOLV~\citep{aydemir2023self}.
SMTC learns to extracts objects from videos by enforcing semantic and instance consistency over time using a student-teacher approach.
SOLV extracts per-frame slots using invariant slot attention~\citep{Biza2023invariant}, applies a temporal consistency module and merges slots using agglomerative clustering; the model is also trained using DINOSAUR-style feature reconstruction, but on masked out intermediate frames.

\section{Method}

\begin{figure}
    \centering
    \includegraphics[width=1.0\textwidth,valign=c]{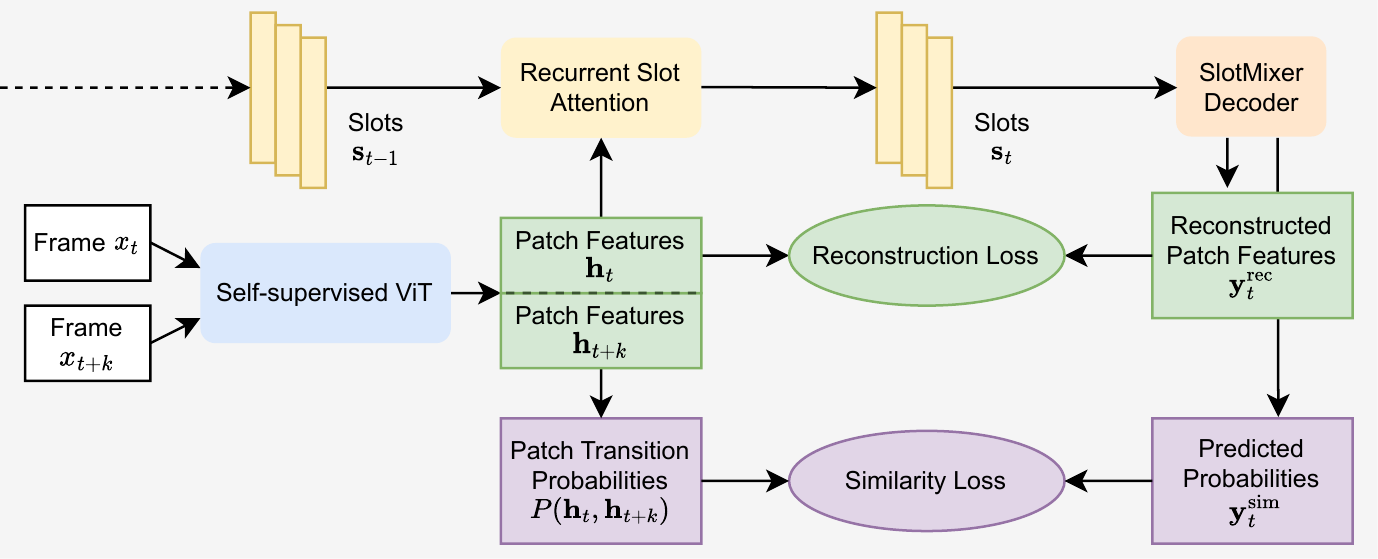}
    \caption{Overview of \method. 
    Object slots $s_t$ are extracted from patch features $\vh_t$ of a self-supervised ViT using time-recurrent slot attention, conditional on slots from the previous time step $t-1$.
    The model is trained by reconstructing the patch features $\vh_t$ of the current frame $x_t$, and by predicting the similarity distribution over patches of a future frame $x_{t+k}$ (see also \fig{fig:feature-sim-loss}).
    The predictions $\vy_t^\text{rec}$ and $\vy_t^{\text{sim}}$ are decoded efficiently using SlotMixer decoder.
    }
    \label{fig:method}
\end{figure}

In this section, we describe the main new components of \method{} --- our proposed object-centric video model --- and its training: a pre-trained self-supervised ViT encoder extracting frame features (\sect{subsec:method-dense-repr}), a temporal similarity loss that adds a motion bias to object discovery (\sect{subsec:method-similarity-loss}), and the SlotMixer decoder to achieve efficient video processing (\sect{subsec:method-slotmixer}).
See \fig{fig:method} for an overview.

\subsection{Slot Attention for Videos with Dense Self-Supervised Representations}
\label{subsec:method-dense-repr}

\method{} is based on the modular video object-centric architecture recently proposed by SAVi~\citep{kipf2022conditional} and also used by STEVE~\citep{Singh2022STEVE}.
Our model has three primary components: (1) a pre-trained self-supervised ViT feature encoder, (2) a recurrent grouping module for temporal slot updates, and (3) the \emph{SlotMixer} decoder (detailed below in \sect{subsec:method-slotmixer}).

We start by processing video frames $\vx_t$, with time steps $t \in \{1,\dots T\}$, into patch features $\vh_t$:
\begin{equation}
\vh_t = f_{\phi}(\vx_t), \quad \vh_t \in \mathbb{R}^{L \times D}
\end{equation}

where $f_{\phi}$ is a self-supervised Vision Transformer encoder (ViT)~\citep{Dosovitskiy2021ViT} with pre-trained parameters $\phi$, and $\vx_t$ is the input at time step $t$.
The ViT encoder processes the image by splitting it to $L$ non-overlapping patches of fixed size (e.g. $16\times16$ pixels), adding positional encoding, and transforming them into $L$ feature vectors $\vh_t$ (see \app{app:vit} for more details on ViTs).
Note that the $i$'th feature retains an association to the $i$'th image patch; the features thus can be spatially arranged.
Next, we transform the features from the encoder with a slot attention module~\citep{Locatello2020SlotAttention} to obtain a latent set $\vs_t = \{\vs_t^{i}\}_{i=1}^{K}$, $s_t^i \in \mathbb R^{M}$ with $K$ slot representations:
\begin{equation}
    \vs_t = \SA_\theta(\vh_t, \vs_{t-1}).
\end{equation}

Slot attention is recurrently initialized with the slots of the previous time step $t-1$, with initial slots $s_0$ sampled independently from a Gaussian distribution with learned location and scale.
Slot attention works by grouping input features into slots by iterating competitive attention steps; we refer to \citet{Locatello2020SlotAttention} for more details.
To train the model, we use a SlotMixer decoder $g_{\paramsdec}$ (see \sect{subsec:method-slotmixer}) to transform the slots $\vs_t$ to outputs $\vy_t = g_{\paramsdec}(\vs_t)$.
Those outputs are used as model predictions for the reconstruction and similarity losses introduced next.

\subsection{Self-Supervised Object Discovery by Predicting Temporal Similarities}
\label{subsec:method-similarity-loss}

We now motivate our novel loss function based on predicting temporal feature similarities.
Video affords the opportunity to discover objects from motion: pixels that consistently move together should be considered as one object, sometimes called the ``common fate'' principle~\citep{Tangemann2023commonfate}.
However, the widely used reconstruction objective --- whether of pixels~\citep{kipf2022conditional}, discrete codes~\citep{Singh2022STEVE} or features~\citep{seitzer2023bridging} --- does not exploit this bias, as to reconstruct the input frame, the changes between frames do not have to be taken into account.

\begin{figure}
    \includegraphics[width=0.49\textwidth,valign=c]{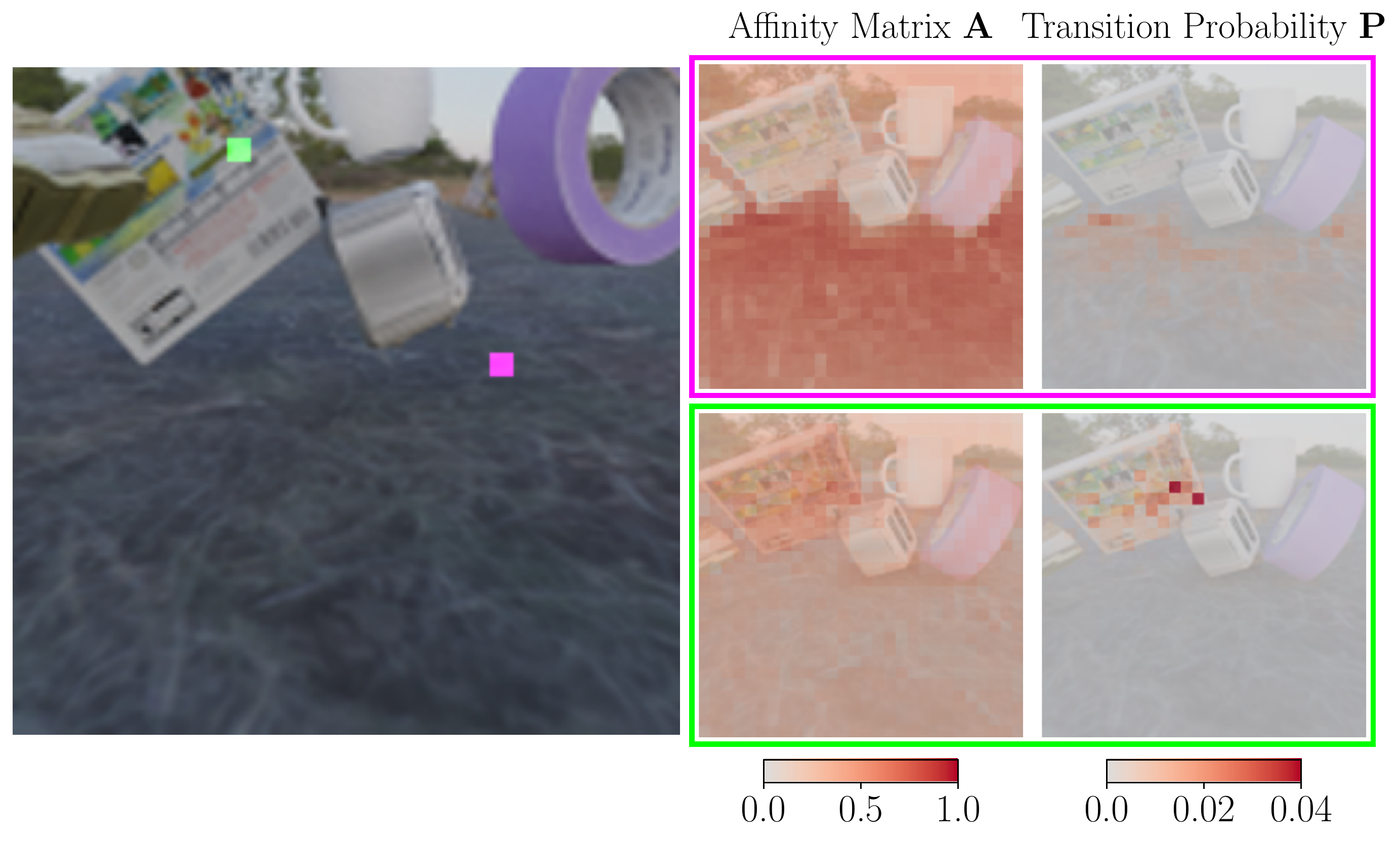}
    \hfill
    \includegraphics[width=0.49\textwidth,valign=c]{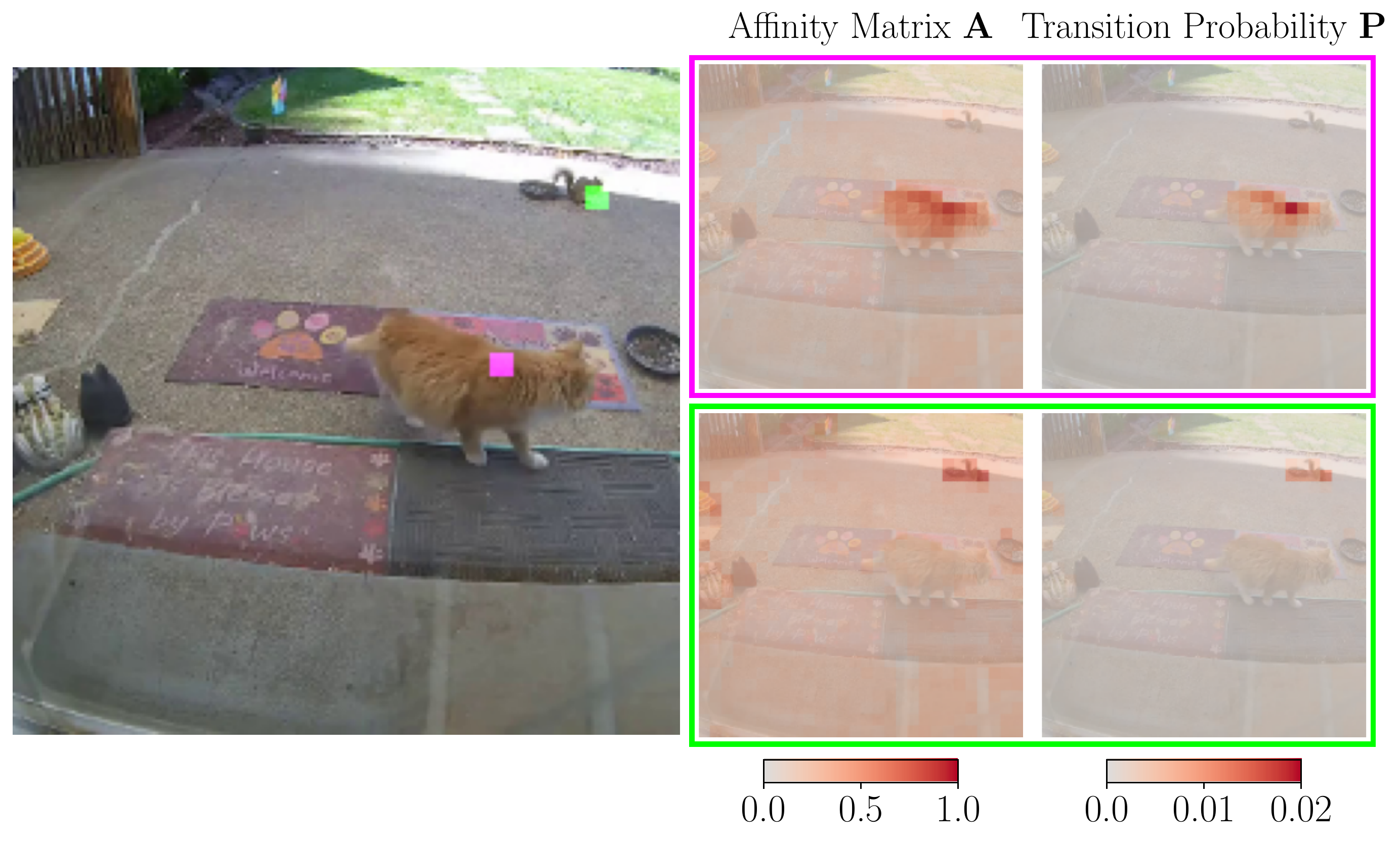}
    \caption{Affinity matrix $\mA_{t, t+k}$ and transition probabilities $\mP_{t, t+k}$ values between patches (marked by purple and green) of the frame $\vx_t$ and patches of the future frame $\vx_{t+k}$ in MOVi-C (left) and  YT-VIS (right). 
    Red indicates maximum affinity/probability. 
    Also see \fig{fig:affinity_additional} for more examples, and \href{https://martius-lab.github.io/videosaur/}{our website} for an interactive visualization of temporal feature similarities.}
    \label{fig:affinity_matrix}
\end{figure}

Taking inspiration from prior work using optical flow as a prediction target~\citep{kipf2022conditional}, we design a self-supervised objective that requires \emph{predicting patch motion}: for each patch, the model needs to predict where all \emph{semantically-similar} patches have moved to $k$ steps into the future.
By comparing self-supervised features describing the patches, we integrate both semantic and motion information; this is in contrast to optical flow prediction, which only relies on motion.
Specifically, we construct an affinity matrix $\mA_{t, t+k}$ with the cosine similarities between all patch features from the present frame $\vh_t$ and all features from some future frame $\vh_{t+k}$:
\begin{equation}
\mA_{t, t+k} = \frac{\vh_t}{\lVert \vh_t \rVert} \cdot \left(\frac{\vh_{t+k}}{\lVert \vh_{t+k} \rVert}\right)^\top, \quad \mA_{t, t+k} \in [-1, 1]^{L \times L}.
\end{equation}
As self-supervised features are highly semantic, the obtained feature similarities are high for patches that share the same semantic interpretation.
Due to the ViT's positional encoding, the similarities also take spatial closeness of patches into account.
\Fig{fig:affinity_matrix} shows several example affinity matrices.

Because there are ambiguities in our similarity-based derivation of feature movements, we frame the prediction task as \emph{modeling a probability distribution} over target patches --- instead of forcing the prediction of an exact target location, like with optical flow prediction.
Thus, we define the probability that patch $i$ moves to patch $j$ by normalizing the rows of the affinity matrix with the softmax, while masking negative similarity values (superscripts refer to the elements of the matrix):
\begin{equation}
\mP^{ij} =
\begin{cases}
\dfrac{\exp(\sfrac{\mA^{ij}}{\tau})}{\sum\limits_{n \in \{j \mid \mA^{ij}\ge0 \}}\hspace*{-1em}\exp(\sfrac{\mA^{in}}{\tau})} & \text{if } \mA^{ij} \ge 0, \\[0.75em]
0 & \text{if } \mA^{ij} < 0,
\end{cases}
\end{equation}
where $\tau$ is the softmax temperature.
The resulting distribution can be interpreted as the \emph{transition probabilities} of a random walk along a graph with image patches as nodes~\citep{jabri2020crw}.
Then, we define the similarity loss as the cross entropy between decoder outputs and transition probabilities:
\begin{equation}
    \mathcal{L}^{\text{sim}}_{\theta, \paramsdec} = \sum_{l=1}^{L} \CE(\mP^{l}_{t, t+k}; \vy_t^{l}).\label{eqn:simloss}
\end{equation}
\Fig{fig:feature-sim-loss} illustrates the loss computation for an example pair of input frames. 

\textbf{Why is this Loss Useful for Object Discovery?\quad}
Predicting which parts of the videos move consistently is most efficient with an object decomposition that captures moving objects. 
This is similar to previous losses predicting optical flow~\citep{kipf2022conditional}.
But in contrast, our loss \eqnp{eqn:simloss} also yields a useful signal for grouping when parts of the frame are \emph{not} moving: as feature similarities capture semantic aspects, the task also requires predicting which patches are semantically similar, helping the grouping into objects \eg by distinguishing fore- and background (see \fig{fig:affinity_matrix}).
Optical flow for grouping also has limits when camera motion is introduced; in our experiments, we find that our loss is more robust in such situations.
Methods based on optical flow or motion masks can also struggle with inaccurate flow/motion mask labels --- unlike our method, which does not require such labels.
This is of particular importance for in-the-wild video, where motion estimation is challenging. %

\textbf{Role of Hyperparameters.\quad}
The loss has two hyperparameters: the time shift into the future $k$ and the softmax temperature $\tau$.
The optimal time shift depends on the expected time scales of movements in the modeled videos and should be chosen accordingly.
The temperature $\tau$ controls the concentration of the distribution onto the maximum. %
Thus, it effectively modulates between two different tasks: accurately estimating the patch motion (low $\tau$), and predicting the similarity of each patch to all other patches (high $\tau$).
In particular in scenes with little movement, the latter may be important to maintain a meaningful prediction task.
In our experiments, we find that the best performance is obtained with a balance between the two, showing that both modes are important.

\textbf{Final Loss.\quad}
While the temporal similarity loss yields state-of-the-art performance on synthetic datasets, as shown below, we found that on real-world data, performance can be further improved by adding the feature reconstruction objective as introduced in \citet{seitzer2023bridging}.
We hypothesize this is because the semantic nature of feature reconstruction adds another useful bias for object discovery.
Thus, the final loss is given by:

\begin{equation}
    \mathcal{L}_{\theta, \paramsdec} = \sum_{t=1}^{T-k} \mathcal{L}_{\theta, \paramsdec}^{\text{sim}}(\mP_{t, t+k}, \vy_t^{\text{sim}}) + \alpha \mathcal{L}_{\theta, \paramsdec}^{\text{rec}}(\vh_t, \vy_t^{\text{rec}}),
\end{equation}

where $\vy_t = [\vy_t^{\text{sim}} \in \mathbb{R}^{L\times L}, \vy_t^{\text{rec}} \in \mathbb{R}^{L \times D}]$ is the output of the SlotMixer decoder $g_{\paramsdec}$ and $\alpha$ is a weighting factor used to make the scales of the two losses similar
(we use a fixed value of $\alpha=0.1$ for all experiments on real-world datasets).
Like in \citet{seitzer2023bridging}, we do not train the ViT encoder $f_\phi$.

\subsection{Efficient Video Object-Centric Learning with the SlotMixer Decoder}
\label{subsec:method-slotmixer}

In video models, resource efficiency is of particular concern:
recurrent frame processing increases the load on compute and memory. %
The standard mixture-based decoder design~\citep{Locatello2020SlotAttention} decodes each output $K$-times, where $K$ is the number of slots, and thus scales linearly with $K$ both in compute and memory.
This can become prohibitive even for a moderate number of slots.
The recently introduced SlotMixer decoder~\citep{sajjadi2022object} for 3D object-centric learning instead has, for all practical purposes, constant overhead in the number of slots, by only decoding once per output.
Thus, we propose to use a SlotMixer decoder $g_{\paramsdec}$ for predicting the probabilities $ \mP_{t, t+k}$ from the slots $\vs_t$.
To adapt the decoder from 3D to 2D outputs, we change the conditioning on 3D query rays to $L$ learned positional embeddings, corresponding to $L$ patch outputs $\vy_t^l$.
See \app{app:mixer_decoder} for more details on the SlotMixer module.

As a consequence of the increased efficiency of SlotMixer, there also is increased flexibility of how slots can be combined to form the outputs. 
Because of this, this decoder has a weaker inductive bias towards object-based groupings compared to the standard mixture-based decoder.
With the standard reconstruction loss, we observed that this manifests in training runs in which no object groupings are discovered.
But in combination with our temporal similarity loss, these instabilities disappear (see \app{app:mixer_stability}).
We attribute this to the \emph{self-supervised nature} of the similarity loss\footnote{Novel-view synthesis, the original task for which SlotMixer was proposed, is similarly a self-supervised prediction task. This may have contributed to the success of SlotMixer in that setting.}; having to predict information that is not directly contained in the input increases the difficulty of the task, reducing the viability of non-object based groupings.

\section{Experiments}

\begin{figure}[t]
  \centering
  \includegraphics[width=\textwidth]{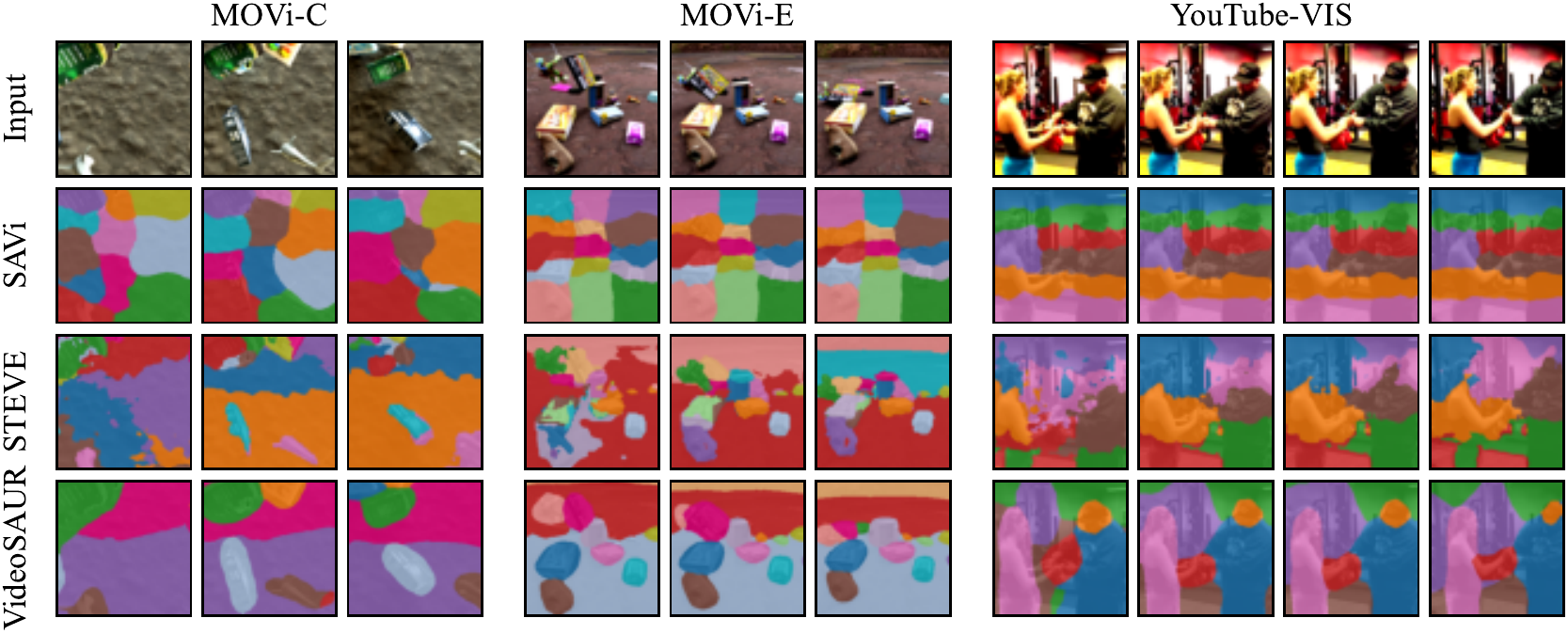}
    \caption{Example predictions of \method{} compared to recent video object-centric methods.}
    \label{fig:combined_example}
\end{figure}

We have conducted a number of experiments to answer the following questions: 
(1) Can object-centric representations be learned from a large number of diverse real-world videos? 
(2) How does \method perform in comparison to other methods on well-established realistic synthetic datasets? 
(3) What are the effects of our proposed temporal feature similarity loss and its parameters? 
(4) Can we transfer the learned object-grouping to unseen datasets? 
(5) How efficient is the SlotMixer decoder in contrast to the mixture-based decoder? %

\subsection{Experimental Setup}
\paragraph{Datasets}
To investigate the characteristics of our proposed method, we utilize three synthetic datasets and three real-world datasets.
For synthetic datasets, we selected the MOVi-C, MOVi-D and MOVi-E datasets~\citep{Greff2021Kubric} that consist of numerous moving objects on complex backgrounds.
Additionally, we evaluate the performance of our method on the challenging YouTube Video Instance Segmentation (YT-VIS) 2021 dataset~\citep{vis2021} as an unconstrained real-world dataset.
Furthermore, we examine how well our model performs when transferred from YT-VIS 2021 to YT-VIS 2019~\citep{vis2019} and DAVIS~\citep{davis2017}.
Finally, we use the COCO dataset~\citep{Lin2014COCO} to study our proposed similarity loss function with image-based object-centric learning.\looseness-1

\paragraph{Metrics}
We evaluate our approach in terms of the quality of the discovered slot masks (output by the decoder), using two metrics:  video foreground ARI (FG-ARI)~\citep{greff2019multi} and video mean best overlap (mBO)~\citep{Arbelaez2014MCG}.
FG-ARI is a video version of a widely used metric in the object-centric literature that measures the similarity of the discovered objects masks to ground truth masks.
This metric mainly measures \emph{how well objects are split}.
mBO assesses the correspondence of the predicted and the ground truth masks using the intersection-over-union (IoU) measure.
In particular, each ground truth mask is matched to the predicted mask with the highest IoU, and the average IoU is then computed across all assigned pairs.
Unlike FG-ARI, mBO also considers background pixels, and provides a measure of \emph{how accurately the masks fit the objects}.
Both metrics also consider the consistency of the assigned object masks over the whole video.

In addition, we also use image-based versions of those metrics~(\textit{Image FG-ARI} and \textit{Image mBO}, computed on individual frames) for comparing with image-based methods. %

\paragraph{Baselines}
We compare our method with two recently proposed methods for unsupervised object-centric learning for videos: SAVi~\citep{kipf2022conditional} and STEVE~\citep{Singh2022STEVE}.
SAVi uses a mixture-based decoder and is trained with image reconstruction.
We use the unconditional version of SAVi.
STEVE uses a transformer decoder and is trained by reconstructing discrete codes of a dVAE~\citep{rolfe2017discrete}.
Similar to \citet{seitzer2023bridging}, we also add a regular block pattern baseline, corresponding to splitting the video into regular block masks of similar size that do not change over time.
By showing the metric values for a trivial decomposition of the video, this baseline is useful to contextualize the results of the other methods.
In addition to video-based methods, we compare our model to image-based methods, including DINOSAUR~\citep{seitzer2023bridging}, LSD~\citep{jiang2023object} and Slot Diffusion~\citep{wu2023slotdiffusion}, showing that our approach performs well in both object separation and mask sharpness.
Last, we also compare our model to two concurrent works discovering objects from real-world video, SMTC~\cite{qian2023semantics} and SOLV~\citep{aydemir2023self}.

\subsection{Comparison with State-of-the-Art Object-Centric Learning Methods}
When comparing \method to STEVE and SAVi, it is evident that \method outperforms the baselines by a significant margin, both in terms of FG-ARI and mBO (see \tab{tab:combined} and \fig{fig:combined_example}). 
On the most challenging synthetic dataset (MOVi-E), \method reaches $73.9$ FG-ARI.
Notably, for the challenging YT-VIS 2021 dataset, both baselines perform comparable or worse than the block pattern baseline in terms of FG-ARI, showing that previous methods struggle to decompose real-world videos into consistent objects.
We additionally compare \method to image-based methods in \app{app:image_comparison}, including strong recent methods (LSD, SlotDiffusion and DINOSAUR), and find that our approach also outperforms the prior image-based SoTA.
Finally, in \app{app:concurrent_work_comparison}, we find that our method performs competitively with concurrent work.

\begin{table}[tb]
\centering
\caption{Comparison with state-of-the-art methods on the MOVi-C, MOVi-E, and YT-VIS datasets. We report foreground adjusted rand index (FG-ARI) and mean best overlap (mBO) over 5 random seeds. 
Both metrics are computed for the whole video (24 frames for MOVi, 6 frames for YT-VIS).
}
\label{tab:combined}
\setlength{\tabcolsep}{5pt}
\begin{tabular}{lrrrrrrrr}\toprule
&\multicolumn{2}{c}{MOVi-C} && \multicolumn{2}{c}{MOVi-E} && \multicolumn{2}{c}{YT-VIS} \\\cmidrule{2-3}\cmidrule{5-6}\cmidrule{8-9}
&FG-ARI & mBO &&FG-ARI & mBO &&FG-ARI & mBO  \\\midrule
Block Pattern &24.2 &11.1 &&36.0 &16.5 &&24 &14.9 \\
SAVi ~\citep{kipf2022conditional} &22.2 ± 2.1 & 13.6 ± 1.6 && 42.8 ± 0.9 & 16.0 ± 0.3&& 11.1 ± 5.6 & 12.7 ± 2.3 \\
STEVE~\citep{Singh2022STEVE} &36.1 ± 2.3&	26.5 ± 1.1 && 50.6 ± 1.7 & 26.6 ± 0.9 && 20.0 ± 1.5	& 20.9 ± 0.5 \\
    \method & \textbf{64.8 ± 1.2}	& \textbf{38.9 ± 0.6} && \textbf{73.9 ± 1.1} & \textbf{35.6 ± 0.5} && \textbf{39.5 ± 0.6}	& \textbf{29.1 ± 0.4} \\
\bottomrule
\end{tabular}
\end{table}

\begin{figure}[tb]
    \centering
    \includegraphics[width=\textwidth,valign=c]{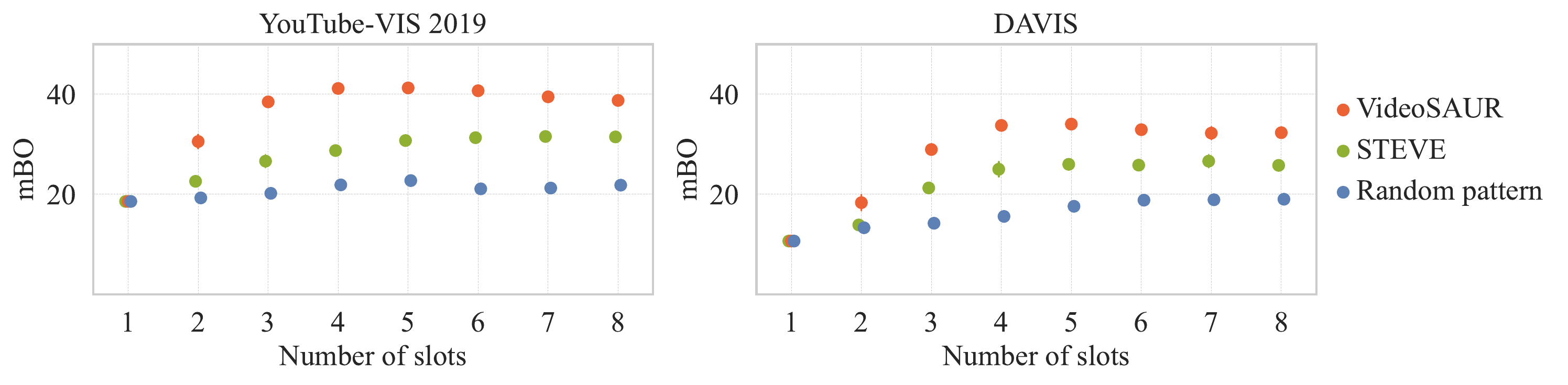}
    \caption{Zero-shot transfer of learned object-centric representations on YT-VIS 2021 to the YT-VIS 2019 and DAVIS datasets for different number of slots.}
    \label{fig:transfer}
\end{figure}

Next, we report how well our method performs in terms of zero-shot transfer to other datasets to show that the learned object discovery does generalize to unseen data. 
In particular, we train \method on the YT-VIS 2021 dataset and evaluate it on the YT-VIS 2019 and DAVIS datasets. 
YT-VIS 2019 has similar object categories, but a smaller number of objects per image.
The DAVIS dataset consists of videos from a fully different distribution than YT-VIS 2021. 
As the number of slots can be changed during evaluation, we test \method with different number of slots, revealing that the optimal number of slots is indeed smaller for these datasets. 
We find that our method achieves a performance of $41.3 \pm 0.9$ mBO on YT-VIS 2019 dataset and $34.0 \pm 0.4$ mBO on DAVIS dataset (see \fig{fig:transfer}), illustrating its capability to effectively transfer the learned representations to previously unseen data with different object categories and numbers of objects.

\paragraph{Long-term Video Consistency}
In addition to studying how \method performs on relatively short 6-frame video segments from YT-VIS, we also evaluate our method on longer videos. 
In \app{app:video_length}, we show the performance for 12-frame and full YT-VIS videos. 
While, as can be expected, performance on longer video segments is smaller in terms of FG-ARI, we show that the gap between \method and the baselines is large, indicating that \method can track the main objects in videos over longer time intervals. 
Closing the gap between short-term and long-term consistency using memory modules~\cite{traub2023learning, gumbsch2021sparsely} is an interesting future direction that could be useful for video prediction~\citep{wu2023slotformer} as well as for object-centric goal-based~\citep{zadaianchuk2020self, mambelli2022compositional} and model-based~\citep{feng2023learning} reinforcement learning.\looseness=-1

\subsection{Analysis}
In this section, we analyze various aspects of our approach, including the importance of the similarity loss, the impact of hyperparameters (time-shift $k$ and softmax temperature $\tau$), and the effect of the choice of self-supervised features and decoder.

\paragraph[Choice of Loss Function]{Choice of Loss Function (\tab{tab:loss_sim_alation_movic} and \tab{tab:loss_sim_alation_ytvis})}
We conduct an ablation study to demonstrate the importance of the proposed temporal similarity loss, comparing and combining it with the feature reconstruction loss~\citep{seitzer2023bridging}.
We also consider predicting the features of the \emph{next frame} (see \app{app:ffrec} for implementation details).
For all datasets, feature reconstruction alone performs significantly worse than the combination of feature reconstruction and temporal similarity loss. 
Predicting the features of the next frame in addition to feature reconstruction also yields improved performance, but is worse than the temporal similarity, suggesting that the success of our loss can be partially explained by the integration of temporal information through future prediction.
Interestingly, on {MOVi-C}, using the temporal similarity loss alone significantly improves the performance over feature reconstruction ($+20$ FG-ARI, $+7$ mBO).
To provide insight into the qualitative differences between the losses, we analyze the videos with the most significant differences in FG-ARI (see 
\fig{fig:loss_diff}):
unlike feature reconstruction, the temporal similarity loss does not fragment the background or large objects into numerous slots, and it exhibits improved object-tracking capabilities even when object size changes. 
To gain further insights, we also consider (ground truth) \emph{optical flow} as a prediction target that only captures motion, but no semantic information (see \app{app:optical_flow} for a detailed discussion).
We find that only predicting optical flow is not enough for a successful scene decomposition, underscoring the importance of integrating both motion and semantic information for real-world object discovery.

\begin{figure}
\begin{minipage}[t]{0.488\textwidth}
        \centering
       \captionof{table}{Loss ablation on MOVi-C.}~\label{tab:loss_sim_alation_movic}
        \footnotesize
        \setlength{\tabcolsep}{2.2pt}
    \begin{tabular}{@{}ccc|rrr@{}}
      \multicolumn{3}{c}{Loss Type} & \multicolumn{2}{c}{Metric} \\
      \toprule
      \multirow{2}{*}{Feat.~Rec.} & Next Frame & \multirow{2}{*}{Temp.~Sim.} & \multirow{2}{*}{FG-ARI} & \multirow{2}{*}{mBO} \\
      & Feat. Pred. & & & \\
      \midrule
      \checkmark & & &40.2 &23.5 \\
      \checkmark &\checkmark & &47.2 &24.7 \\
            & &\checkmark &\textbf{60.8} &\textbf{30.5} \\
      \checkmark & &\checkmark &60.7 &30.3 \\

      \bottomrule
    \end{tabular}
\end{minipage}
\hfill
\begin{minipage}[t]{0.488\textwidth}
        \centering
        \captionof{table}{Loss ablation on YT-VIS.}~\label{tab:loss_sim_alation_ytvis}
        \footnotesize
        \setlength{\tabcolsep}{2.2pt}
    \begin{tabular}{@{}ccc|rrr@{}}
      \multicolumn{3}{c}{Loss Type} & \multicolumn{2}{c}{Metric} \\
      \toprule
      \multirow{2}{*}{Feat.~Rec.} & Next Frame & \multirow{2}{*}{Temp.~Sim.} & \multirow{2}{*}{FG-ARI} & \multirow{2}{*}{mBO} \\
      & Feat. Pred. & & & \\
      \midrule
      \checkmark & & &35.4 &26.7 \\
      \checkmark &\checkmark & &37.9 &27.3 \\
            & &\checkmark &26.2 & \textbf{29.1} \\
      \checkmark & &\checkmark &\textbf{39.5 }&\textbf{29.1} \\
      \bottomrule
    \end{tabular}
\end{minipage}
\end{figure}

\paragraph{Robustness to Camera Motion (\Tab{tab:camera_motion})}
Next, we investigate if \method training with the similarity loss is robust to camera motion, as such motion makes isolating the object motion more difficult. 
As a controlled experiment, we compare between MOVi-D (without camera motion) and MOVi-E (with camera motion), and train \method using only the temporal similarity loss.
We contrast with SAVi trained with optical flow prediction\footnote{SAVi results with optical flow are from \citet{Greff2021Kubric}.}, and find that \method is more robust to camera motion, performing better on the MOVi-E dataset than on the MOVi-D dataset ($+6.8$ vs $-16.7$ FG-ARI for SAVi).\looseness-1

\paragraph{Choice of Decoder (\Tab{tab:decoder})} 
We analyze how our method performs with different decoders and find that both the MLP broadcast decoder~\citep{seitzer2023bridging} and our proposed SlotMixer decoder can be used for optimizing the temporal similarity loss. 
\method with the MLP broadcast decoder achieves similar performance on YT-VIS and MOVi datasets, but requires 2--3 times more GPU memory (see ~\app{app:mixer_decoder} for more details and ~\Tab{tab:model_comparison_movi_e} for the detailed comparison of decoders on MOVI-E dataset).
Thus, we suggest to use the SlotMixer decoder for efficient video processing.\looseness-1

\begin{table}
    \begin{minipage}[t]{0.44\textwidth}
       \centering
       \setlength{\tabcolsep}{2pt}
       \captionof{table}{Robustness to introducing camera motion (MOVi-D $\rightarrow$ MOVi-E).%
       }\label{tab:camera_motion}
        \small
    \begin{tabular}{@{}crr@{}}
      \toprule
       & MOVi-D & MOVi-E \\
      \midrule
      SAVi (optical flow)~\cite{Greff2021Kubric} & 19.4 & 2.7 \\
      \method (temporal sim.) & 55.7 & 62.5 \\
      \bottomrule
    \end{tabular}
    \end{minipage}\hfill
    \begin{minipage}[t]{0.54\textwidth}
        \centering
        \captionof{table}{
        Decoder comparison on MOVi-C and YT-VIS.
        }\label{tab:decoder}
        \footnotesize
        \setlength{\tabcolsep}{0.25em}
        \begin{tabular}{@{}lrrrrrr@{}}\toprule
        &\multicolumn{2}{c}{MOVi-C} && \multicolumn{2}{c}{YT-VIS} & Memory \\\cmidrule{2-4}\cmidrule{5-7}
        &FG-ARI & mBO && FG-ARI & mBO & GB @24 slots\\
        \midrule
        Mixer & 60.8 &30.5 && 39.5& 29.1 & 24\\
        MLP   & 64.2 & 27.2 &&  39.0 & 29.1 & 70\\
    \bottomrule
    \end{tabular}
    \end{minipage}
\end{table}

\begin{figure}
  \centering
  \begin{subfigure}[b]{0.34\textwidth}
    \includegraphics[width=\textwidth]{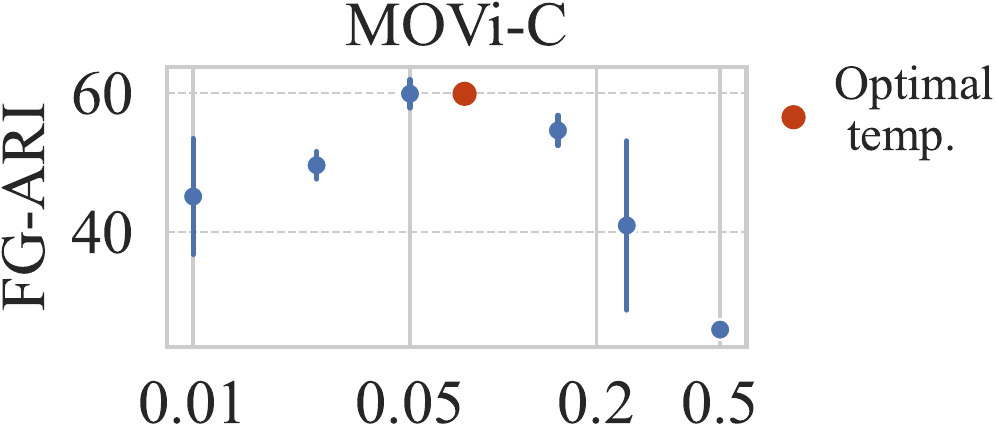}
    \vspace{-1.0em}
    \caption{Softmax temperature $\tau$.}
    \label{fig:softmax_temperature}
  \end{subfigure}
      \hfill
  \begin{subfigure}[b]{0.63\textwidth}
    \includegraphics[height=2.9cm]{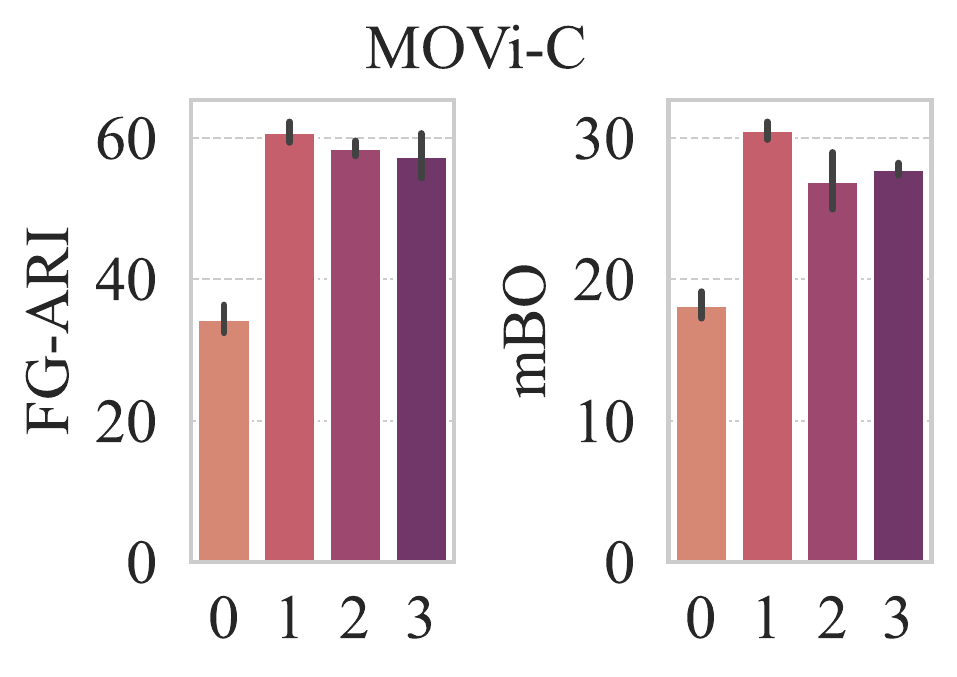}
    \includegraphics[height=2.9cm]{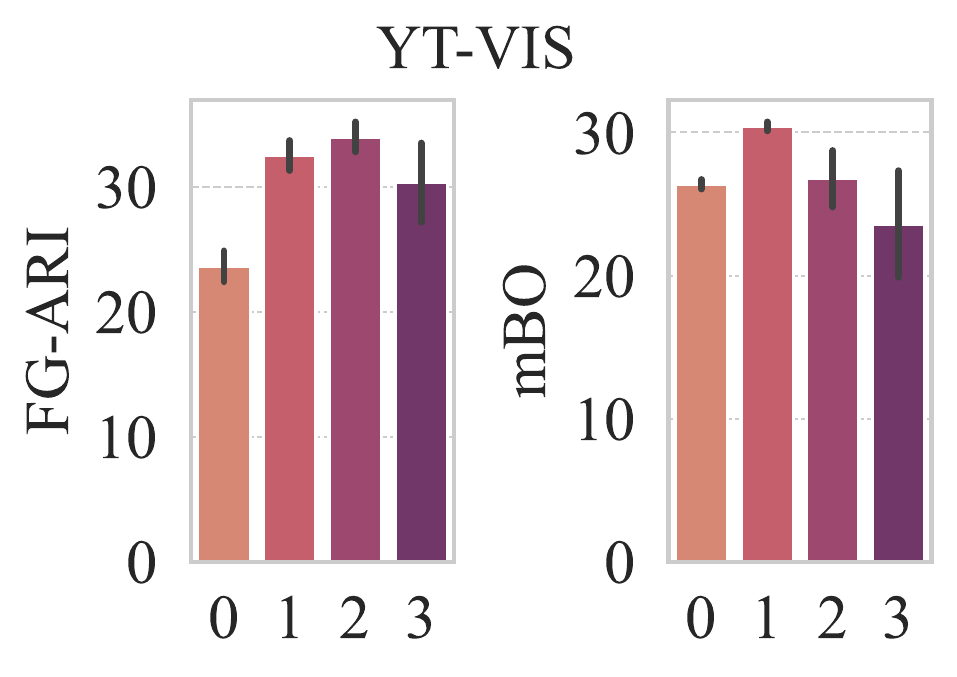}
    \vspace{-0.5em}
    \caption{Target time-shift $k$ on MOVi-C and YT-VIS datasets.}
    \label{fig:time_shift}
  \end{subfigure}

  \vspace{0.5em}
  \begin{subfigure}[b]{0.42\textwidth}
    \includegraphics[width=\linewidth]{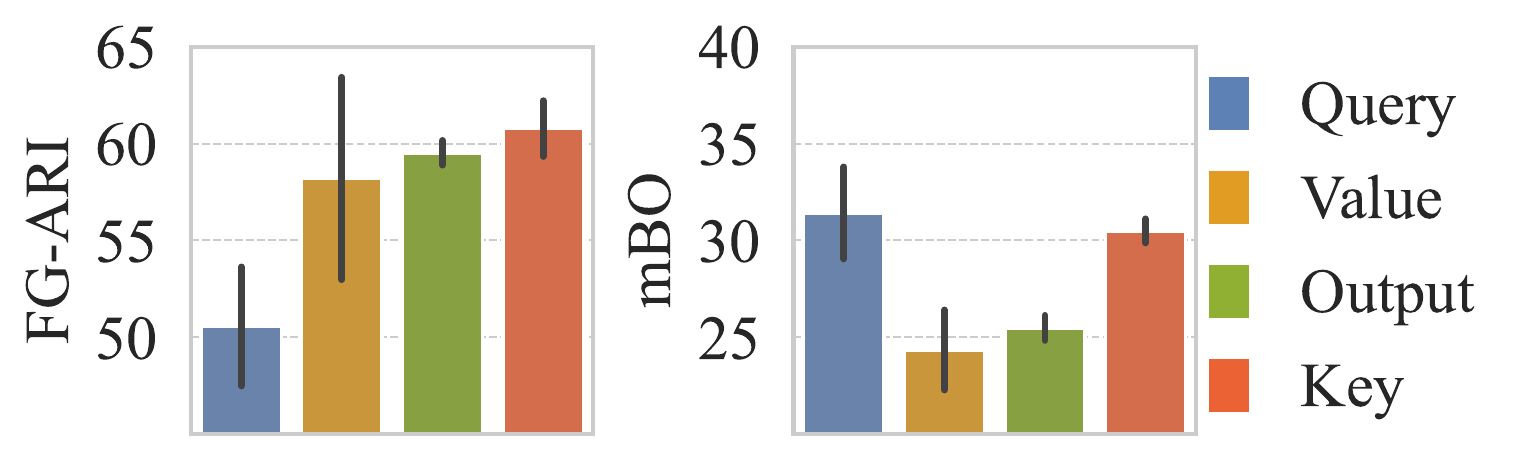}
    \vspace{-1.5em}
    \caption{DINO ViT features.
    }
    \label{fig:keys}
  \end{subfigure} \hfill
    \begin{subfigure}[b]{0.56\textwidth}
    \includegraphics[width=\linewidth]{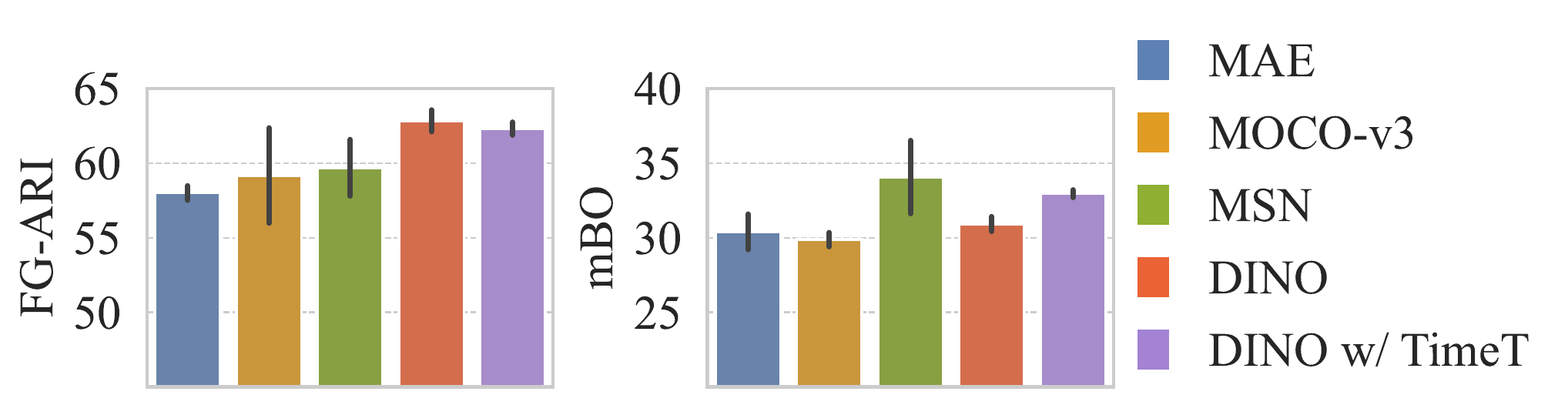}
    \vspace{-1.5em}
    \caption{Self-supervised representation method.}\label{fig:features_choice}
  \end{subfigure}%
  \caption{Studying the effect of different parameters of the temporal similarity loss. }
  \label{fig:ablation_params}
\end{figure}

\paragraph{Softmax Temperature (\Fig{fig:softmax_temperature})}
We train \method with DINO S/16 features using different softmax temperatures $\tau$.
We find that there is a sweet spot in terms of grouping performance at $\tau=0.075$.
Lower and higher temperatures lead to high variance across seeds, potentially because there is not enough training signal with very peaked (low $\tau$) and diffuse (high $\tau$) target distributions.

\paragraph{Target Time-shift (\Fig{fig:time_shift})}
We train \method with DINO S/16 features using different time-shifts $k$ to construct the affinity matrix $\mA_{t, t+k}$. 
On both synthetic and real-world datasets, $k=1$ generally performs best. 
Interestingly, we find that for $k=0$, performance drops, indicating that predicting pure self-similarities is not a sufficient task for discovering objects on its own.

\paragraph{Choices for Self-Supervised Features (\Figs{fig:keys}{fig:features_choice})}
We study two questions about the usage of the ViT features: which ViT features (queries/keys/values/outputs) should be used for the temporal similarity loss? 
Do different self-supervised representations result in different performance?
In \fig{fig:keys}, we observe that using DINO ``key'' and ``query'' features leads to significantly larger mBO, while for FG-ARI ``query'' is worse and the other features are similar.
Potentially, this is because keys are used in the ViT's self-attention and thus could be particularly good to compare with the scalar product similarity. 
Consequently, \method uses ``key'' features in all other experiments.
Moreover, we study if the temporal similarity loss is compatible with different self-supervised representations. 
In \fig{fig:features_choice}, we show that \method works well with 4 different types of representations, with MSN~\cite{Assran2022MSN} and DINO~\citep{Caron2021DINO} performing slightly better than MAE~\citep{He2022MAE} and MOCO-v3~\citep{Chen2021MoCov3}. 
We also demonstrate that \emph{further fine-tuning the DINO features} utilizing a self-supervised temporal-alignment clustering approach named TimeTuning~\citep{Salehi_2023_ICCV} on unlabeled videos enhances the mask quality of \method.\looseness-1

\paragraph{Pre-training Dataset~(\Tab{tab:pretraining})}
All self-supervised methods we utilize are trained on the ImageNet dataset, which a) has a strong bias towards object-centricness as its images mostly contain single objects, and b) introduces a large number of additional images external to the dataset we are training \method on.
An interesting question is whether a) and b) are actually required for the success of our method. 
To answer it, we train a ViT-B/16 encoder from scratch on the MOVi-E dataset using the MAE method, and then train \method using the obtained features.
Interestingly, we find that the features from MOVi-E yield similar results compared to ImageNet-trained features (although with slight drops in mask quality), demonstrating that \method is able to perform high-quality object discovery even without access to external data.
This result also has broader implications as it potentially increases the applicability of feature reconstruction-based object-centric methods to datasets fully out of the domain of ImageNet.
It also raises a follow-up question: what properties of the pre-training dataset (and method) are important to obtain good target features for object discovery?

\begin{table}
    \centering
    \caption{Comparing \method with features trained on MOVi-E (\emph{MAE+MOVi-E}) to features trained on ImageNet (\emph{MAE+ImageNet}).
    For MAE+MOVi-E, we pre-train a ViT-B/16 using the self-supervised MAE method on MOVi-E for 200 epochs.
    \method is able to perform high-quality object discovery even without access to any external data.
    }
    \label{tab:pretraining}
    \setlength{\tabcolsep}{5pt}
    \small
    \begin{tabular}{lrrrrrr}\toprule
     &\multicolumn{2}{c}{MOVi-C} && \multicolumn{2}{c}{MOVi-E} \\\cmidrule{2-3}\cmidrule{5-6}
    &FG-ARI & mBO && FG-ARI & mBO   \\\midrule
    VideoSAUR w/ \emph{MAE+ImageNet} features & 58.0 & 30.4 && 72.8  & 27.1  \\
    VideoSAUR w/ \emph{MAE+MOVi-E} features & 59.8 & 27.5 && 70.6  & 23.3  \\
    \bottomrule
    \end{tabular}
\end{table}

\section{Conclusion}
This paper presents the first method for unsupervised video-based object-centric learning that scales to diverse, unconstrained real-world datasets such as YouTube-VIS. 
By leveraging dense self-supervised features and extracting motion information with temporal similarity loss, we demonstrate superior performance on both synthetic and real-world video datasets.
We hope our new loss function can inspire the design of further self-supervised losses for object-centric learning, especially in the video domain where natural self-supervision is available.

Still, our method does not come without limitations:
in longer videos with occlusions, slots can get reassigned to different objects or the background (see \fig{fig:ytvis2021_fal_case} for visualizations of failure cases).
\method also inherits a limitation of all slot attention-based method, namely that the the number of slots is static and needs to be chosen a priori.
Similar to DINOSAUR~\citep{seitzer2023bridging}, the quality of the object masks is restricted by the patch-based nature of the decoder.
Finally, while the datasets we use in this work are significantly less constrained compared to datasets used by prior work, they still do not capture the full open-world setting that object-centric learning aspires to solve. %
Overcoming these limitations is a great direction for future work.%

\clearpage
\section*{Acknowledgements} We thank Martin Butz, Cansu Sancaktar and Manuel Traub for useful suggestions and discussions. 
The authors thank the International Max Planck Research School for Intelligent Systems (IMPRS-IS) for supporting Maximilian Seitzer. Andrii Zadaianchuk is supported by the Max Planck ETH Center for Learning Systems. Georg Martius is a member of the Machine Learning Cluster of Excellence, funded by the Deutsche Forschungsgemeinschaft (DFG, German Research Foundation) under Germany’s Excellence Strategy – EXC number 2064/1 – Project number 390727645. We acknowledge the support from the German Federal Ministry of Education and Research (BMBF) through the Tübingen AI Center (FKZ: 01IS18039B).

\bibliography{main}

\begin{thebibliography}{59}
\providecommand{\natexlab}[1]{#1}
\providecommand{\url}[1]{\texttt{#1}}
\expandafter\ifx\csname urlstyle\endcsname\relax
  \providecommand{\doi}[1]{doi: #1}\else
  \providecommand{\doi}{doi: \begingroup \urlstyle{rm}\Url}\fi

\bibitem[Burgess et~al.(2019)Burgess, Matthey, Watters, Kabra, Higgins, Botvinick, and Lerchner]{burgess2019monet}
Christopher~P. Burgess, Loic Matthey, Nicholas Watters, Rishabh Kabra, Irina Higgins, Matt Botvinick, and Alexander Lerchner.
\newblock {MONet: Unsupervised Scene Decomposition and Representation}.
\newblock \emph{arXiv:1901.11390}, 2019.
\newblock URL \url{https://arxiv.org/abs/1901.11390}.

\bibitem[Greff et~al.(2019)Greff, Kaufman, Kabra, Watters, Burgess, Zoran, Matthey, Botvinick, and Lerchner]{greff2019multi}
Klaus Greff, Rapha{\"e}l~Lopez Kaufman, Rishabh Kabra, Nick Watters, Christopher Burgess, Daniel Zoran, Loic Matthey, Matthew Botvinick, and Alexander Lerchner.
\newblock {Multi-Object Representation Learning with Iterative Variational Inference}.
\newblock In \emph{ICML}, 2019.
\newblock URL \url{https://arxiv.org/abs/1903.00450}.

\bibitem[Locatello et~al.(2020)Locatello, Weissenborn, Unterthiner, Mahendran, Heigold, Uszkoreit, Dosovitskiy, and Kipf]{Locatello2020SlotAttention}
Francesco Locatello, Dirk Weissenborn, Thomas Unterthiner, Aravindh Mahendran, Georg Heigold, Jakob Uszkoreit, Alexey Dosovitskiy, and Thomas Kipf.
\newblock {Object-Centric Learning with Slot Attention}.
\newblock In \emph{NeurIPS}, 2020.
\newblock URL \url{https://proceedings.neurips.cc/paper/2020/file/8511df98c02ab60aea1b2356c013bc0f-Paper.pdf}.

\bibitem[Jiang et~al.(2020)Jiang, Janghorbani, {de Melo}, and Ahn]{Jiang2020SCALOR}
Jindong Jiang, Sepehr Janghorbani, Gerard {de Melo}, and Sungjin Ahn.
\newblock {SCALOR: Generative World Models with Scalable Object Representations}.
\newblock In \emph{ICLR}, 2020.
\newblock URL \url{https://openreview.net/pdf?id=SJxrKgStDH}.

\bibitem[Kipf et~al.(2022)Kipf, Elsayed, Mahendran, Stone, Sabour, Heigold, Jonschkowski, Dosovitskiy, and Greff]{kipf2022conditional}
Thomas Kipf, Gamaleldin~Fathy Elsayed, Aravindh Mahendran, Austin Stone, Sara Sabour, Georg Heigold, Rico Jonschkowski, Alexey Dosovitskiy, and Klaus Greff.
\newblock {Conditional Object-centric Learning from Video}.
\newblock In \emph{ICLR}, 2022.
\newblock URL \url{https://openreview.net/forum?id=aD7uesX1GF_}.

\bibitem[Elsayed et~al.(2022)Elsayed, Mahendran, van Steenkiste, Greff, Mozer, and Kipf]{elsayed2022savi++}
Gamaleldin~Fathy Elsayed, Aravindh Mahendran, Sjoerd van Steenkiste, Klaus Greff, Michael~Curtis Mozer, and Thomas Kipf.
\newblock {SAVi++: Towards End-to-End Object-Centric Learning from Real-World Videos}.
\newblock In \emph{NeurIPS}, 2022.
\newblock URL \url{https://openreview.net/forum?id=fT9W53lLxNS}.

\bibitem[Singh et~al.(2022{\natexlab{a}})Singh, Wu, and Ahn]{Singh2022STEVE}
Gautam Singh, Yi-Fu Wu, and Sungjin Ahn.
\newblock {Simple Unsupervised Object-Centric Learning for Complex and Naturalistic Videos}.
\newblock In \emph{NeurIPS}, 2022{\natexlab{a}}.
\newblock URL \url{https://openreview.net/forum?id=eYfIM88MTUE}.

\bibitem[Bao et~al.(2022)Bao, Tokmakov, Jabri, Wang, Gaidon, and Hebert]{Bao2022ObjectsThatMove}
Zhipeng Bao, Pavel Tokmakov, Allan Jabri, Yu-Xiong Wang, Adrien Gaidon, and Martial Hebert.
\newblock {Discovering Objects that Can Move}.
\newblock \emph{CVPR}, 2022.
\newblock URL \url{https://arxiv.org/abs/2203.10159}.

\bibitem[Seitzer et~al.(2023)Seitzer, Horn, Zadaianchuk, Zietlow, Xiao, Simon-Gabriel, He, Zhang, Sch{\"o}lkopf, Brox, and Locatello]{seitzer2023bridging}
Maximilian Seitzer, Max Horn, Andrii Zadaianchuk, Dominik Zietlow, Tianjun Xiao, Carl-Johann Simon-Gabriel, Tong He, Zheng Zhang, Bernhard Sch{\"o}lkopf, Thomas Brox, and Francesco Locatello.
\newblock Bridging the gap to real-world object-centric learning.
\newblock In \emph{ICLR}, 2023.
\newblock URL \url{https://openreview.net/forum?id=b9tUk-f_aG}.

\bibitem[Caron et~al.(2021)Caron, Touvron, Misra, Jégou, Mairal, Bojanowski, and Joulin]{Caron2021DINO}
Mathilde Caron, Hugo Touvron, Ishan Misra, Hervé Jégou, Julien Mairal, Piotr Bojanowski, and Armand Joulin.
\newblock {Emerging Properties in Self-Supervised Vision Transformers}.
\newblock \emph{ICCV}, 2021.
\newblock URL \url{https://arxiv.org/abs/2104.14294}.

\bibitem[He et~al.(2022)He, Chen, Xie, Li, Doll\'ar, and Girshick]{He2022MAE}
Kaiming He, Xinlei Chen, Saining Xie, Yanghao Li, Piotr Doll\'ar, and Ross Girshick.
\newblock {Masked Autoencoders are Scalable Vision Learners}.
\newblock In \emph{CVPR}, 2022.
\newblock URL \url{https://arxiv.org/abs/2111.06377}.

\bibitem[Greff et~al.(2022)Greff, Belletti, Beyer, Doersch, Du, Duckworth, Fleet, Gnanapragasam, Golemo, Herrmann, Kipf, Kundu, Lagun, Laradji, Liu, Meyer, Miao, Nowrouzezahrai, Oztireli, Pot, Radwan, Rebain, Sabour, Sajjadi, Sela, Sitzmann, Stone, Sun, Vora, Wang, Wu, Yi, Zhong, and Tagliasacchi]{Greff2021Kubric}
Klaus Greff, Francois Belletti, Lucas Beyer, Carl Doersch, Yilun Du, Daniel Duckworth, David~J. Fleet, Dan Gnanapragasam, Florian Golemo, Charles Herrmann, Thomas Kipf, Abhijit Kundu, Dmitry Lagun, Issam Laradji, Hsueh-Ti~(Derek) Liu, Henning Meyer, Yishu Miao, Derek Nowrouzezahrai, Cengiz Oztireli, Etienne Pot, Noha Radwan, Daniel Rebain, Sara Sabour, Mehdi S.~M. Sajjadi, Matan Sela, Vincent Sitzmann, Austin Stone, Deqing Sun, Suhani Vora, Ziyu Wang, Tianhao Wu, Kwang~Moo Yi, Fangcheng Zhong, and Andrea Tagliasacchi.
\newblock {Kubric: A Scalable Dataset Generator}.
\newblock In \emph{CVPR}, 2022.
\newblock URL \url{https://arxiv.org/abs/2203.03570}.

\bibitem[Sajjadi et~al.(2022)Sajjadi, Duckworth, Mahendran, van Steenkiste, Pavetic, Lucic, Guibas, Greff, and Kipf]{sajjadi2022object}
Mehdi~SM Sajjadi, Daniel Duckworth, Aravindh Mahendran, Sjoerd van Steenkiste, Filip Pavetic, Mario Lucic, Leonidas~J Guibas, Klaus Greff, and Thomas Kipf.
\newblock Object scene representation transformer.
\newblock In \emph{NeurIPS}, 2022.
\newblock URL \url{https://arxiv.org/abs/2206.06922}.

\bibitem[Yang et~al.(2021)Yang, Fan, Fu, and Xu]{vis2021}
Linjie Yang, Yuchen Fan, Yang Fu, and Ning Xu.
\newblock The 3rd large-scale video object segmentation challenge - video instance segmentation track, June 2021.
\newblock URL \url{https://youtube-vos.org/dataset/vis}.

\bibitem[Kosiorek et~al.(2018)Kosiorek, Kim, Teh, and Posner]{Kosiorek2018SQAIR}
Adam Kosiorek, Hyunjik Kim, Yee~Whye Teh, and Ingmar Posner.
\newblock {Sequential Attend, Infer, Repeat: Generative Modelling of Moving Objects}.
\newblock In \emph{NeurIPS}, 2018.
\newblock URL \url{https://proceedings.neurips.cc/paper/2018/hash/7417744a2bac776fabe5a09b21c707a2-Abstract.html}.

\bibitem[Crawford and Pineau(2020)]{crawford2020spair}
Eric Crawford and Joelle Pineau.
\newblock Spatially invariant unsupervised object detection with convolutional neural networks.
\newblock In \emph{AAAI}, 2020.
\newblock URL \url{https://arxiv.org/abs/1911.09033}.

\bibitem[Lin et~al.(2020)Lin, Fu, Peri, Sun, Singh, Deng, Jiang, and Ahn]{Lin2020SPACE}
Zhixuan Lin, Yi-Wu Fu, Skand~Vishwanath Peri, Weihao Sun, Gautam Singh, Fei Deng, Jindong Jiang, and Sungjin Ahn.
\newblock {SPACE: Unsupervised Object-Oriented Scene Representation via Spatial Attention and Decomposition}.
\newblock In \emph{ICLR}, 2020.
\newblock URL \url{https://openreview.net/forum?id=rkl03ySYDH}.

\bibitem[Greff et~al.(2017)Greff, van Steenkiste, and Schmidhuber]{greff2017neural}
Klaus Greff, Sjoerd van Steenkiste, and Jürgen Schmidhuber.
\newblock Neural expectation maximization.
\newblock In \emph{NeurIPS}, 2017.
\newblock URL \url{https://arxiv.org/abs/1708.03498}.

\bibitem[van Steenkiste et~al.(2018)van Steenkiste, Chang, Greff, and Schmidhuber]{vanstennkiste2018relational}
Sjoerd van Steenkiste, Michael Chang, Klaus Greff, and Jürgen Schmidhuber.
\newblock Relational neural expectation maximization: Unsupervised discovery of objects and their interactions.
\newblock In \emph{ICLR}, 2018.
\newblock URL \url{https://openreview.net/forum?id=ryH20GbRW}.

\bibitem[Veerapaneni et~al.(2019)Veerapaneni, Co-Reyes, Chang, Janner, Finn, Wu, Tenenbaum, and Levine]{veerapaneni2019entity}
Rishi Veerapaneni, John~D. Co-Reyes, Michael Chang, Michael Janner, Chelsea Finn, Jiajun Wu, Joshua Tenenbaum, and Sergey Levine.
\newblock {Entity Abstraction in Visual Model-based Reinforcement Learning}.
\newblock In \emph{Conference on Robot Learning}, 2019.
\newblock URL \url{https://arxiv.org/abs/1910.12827}.

\bibitem[Zoran et~al.(2021)Zoran, Kabra, Lerchner, and Rezende]{Zoran2021parts}
Daniel Zoran, Rishabh Kabra, Alexander Lerchner, and Danilo~J. Rezende.
\newblock {PARTS}: Unsupervised segmentation with slots, attention and independence maximization.
\newblock In \emph{ICCV}, 2021.
\newblock URL \url{https://ieeexplore.ieee.org/document/9711314}.

\bibitem[Weis et~al.(2021)Weis, Chitta, Sharma, Brendel, Bethge, Geiger, and Ecker]{weis2021benchmarking}
Marissa~A Weis, Kashyap Chitta, Yash Sharma, Wieland Brendel, Matthias Bethge, Andreas Geiger, and Alexander~S. Ecker.
\newblock {Benchmarking Unsupervised Object Representations for Video Sequences}.
\newblock \emph{JMLR}, 2021.
\newblock URL \url{https://jmlr.org/papers/v22/21-0199.html}.

\bibitem[Kabra et~al.(2021)Kabra, Zoran, Erdogan, Matthey, Creswell, Botvinick, Lerchner, and Burgess]{kabra2021simone}
Rishabh Kabra, Daniel Zoran, Goker Erdogan, Loic Matthey, Antonia Creswell, Matthew Botvinick, Alexander Lerchner, and Christopher~P. Burgess.
\newblock {SIMON}e: View-invariant, temporally-abstracted object representations via unsupervised video decomposition.
\newblock In \emph{NeurIPS}, 2021.
\newblock URL \url{https://openreview.net/forum?id=YSzTMntO1KY}.

\bibitem[Traub et~al.(2023{\natexlab{a}})Traub, Otte, Menge, Karlbauer, Thuemmel, and Butz]{traub2023learning}
Manuel Traub, Sebastian Otte, Tobias Menge, Matthias Karlbauer, Jannik Thuemmel, and Martin~V. Butz.
\newblock {Learning What and Where: Disentangling Location and Identity Tracking Without Supervision}.
\newblock In \emph{ICLR}, 2023{\natexlab{a}}.
\newblock URL \url{https://openreview.net/forum?id=NeDc-Ak-H_}.

\bibitem[Safadoust and G{\"u}ney(2023)]{safadoust2023multi}
Sadra Safadoust and Fatma G{\"u}ney.
\newblock Multi-object discovery by low-dimensional object motion.
\newblock In \emph{ICCV}, 2023.
\newblock URL \url{https://arxiv.org/abs/2307.08027}.

\bibitem[Singh et~al.(2022{\natexlab{b}})Singh, Deng, and Ahn]{Singh2022SLATE}
Gautam Singh, Fei Deng, and Sungjin Ahn.
\newblock {Illiterate DALL-E Learns to Compose}.
\newblock In \emph{ICLR}, 2022{\natexlab{b}}.
\newblock URL \url{https://openreview.net/forum?id=h0OYV0We3oh}.

\bibitem[Karazija et~al.(2021)Karazija, Laina, and Rupprecht]{Karazija2021clevrtex}
Laurynas Karazija, Iro Laina, and Christian Rupprecht.
\newblock {ClevrTex: A Texture-Rich Benchmark for Unsupervised Multi-Object Segmentation}.
\newblock In \emph{NeurIPS Track on Datasets and Benchmarks}, 2021.
\newblock URL \url{https://arxiv.org/abs/2111.10265}.

\bibitem[Geiger et~al.(2013)Geiger, Lenz, Stiller, and Urtasun]{Geiger2013KITTI}
Andreas Geiger, Philip Lenz, Christoph Stiller, and Raquel Urtasun.
\newblock {Vision meets Robotics: The KITTI Dataset}.
\newblock \emph{International Journal of Robotics Research}, 2013.
\newblock URL \url{https://www.cvlibs.net/publications/Geiger2013IJRR.pdf}.

\bibitem[Yang and Yang(2022)]{yang2022promising}
Yafei Yang and Bo~Yang.
\newblock {Promising or Elusive? Unsupervised Object Segmentation from Real-world Single Images}.
\newblock In \emph{NeurIPS}, 2022.
\newblock URL \url{https://openreview.net/forum?id=DzPWTwfby5d}.

\bibitem[Biza et~al.(2023)Biza, van Steenkiste, Sajjadi, Elsayed, Mahendran, and Kipf]{Biza2023invariant}
Ondrej Biza, Sjoerd van Steenkiste, Mehdi S.~M. Sajjadi, Gamaleldin~F. Elsayed, Aravindh Mahendran, and Thomas Kipf.
\newblock Invariant slot attention: Object discovery with slot-centric reference frames.
\newblock In \emph{ICML}, 2023.
\newblock URL \url{https://arxiv.org/abs/2302.04973}.

\bibitem[Kim et~al.(2023)Kim, Choi, Choi, and Kim]{kim2023shepherding}
Jinwoo Kim, Janghyuk Choi, Ho-Jin Choi, and Seon~Joo Kim.
\newblock Shepherding slots to objects: Towards stable and robust object-centric learning.
\newblock In \emph{CVPR}, 2023.
\newblock URL \url{https://arxiv.org/abs/2303.17842}.

\bibitem[Pervez et~al.(2023)Pervez, Lippe, and Gavves]{pervez2023differentiable}
Adeel Pervez, Phillip Lippe, and Efstratios Gavves.
\newblock Differentiable mathematical programming for object-centric representation learning.
\newblock In \emph{ICLR}, 2023.
\newblock URL \url{https://openreview.net/forum?id=1J-ZTr7aypY}.

\bibitem[Engelcke et~al.(2021)Engelcke, Jones, and Posner]{engelcke2021genesis2}
Martin Engelcke, Oiwi~Parker Jones, and Ingmar Posner.
\newblock {GENESIS-V2: Inferring Unordered Object Representations without Iterative Refinement}.
\newblock In \emph{NeurIPS}, 2021.
\newblock URL \url{https://openreview.net/forum?id=nRBZWEUhIhW}.

\bibitem[Chang et~al.(2022)Chang, Griffiths, and Levine]{Chang2022implicitdiff}
Michael Chang, Thomas~L. Griffiths, and Sergey Levine.
\newblock Object representations as fixed points: Training iterative refinement algorithms with implicit differentiation.
\newblock In \emph{NeurIPS}, 2022.
\newblock URL \url{https://arxiv.org/abs/2207.00787}.

\bibitem[Jia et~al.(2023)Jia, Liu, and Huang]{Jia2023boqsa}
Baoxiong Jia, Yu~Liu, and Siyuan Huang.
\newblock Improving object-centric learning with query optimization.
\newblock In \emph{ICLR}, 2023.
\newblock URL \url{https://arxiv.org/abs/2210.08990}.

\bibitem[Jiang et~al.(2023)Jiang, Deng, Singh, and Ahn]{jiang2023object}
Jindong Jiang, Fei Deng, Gautam Singh, and Sungjin Ahn.
\newblock Object-centric slot diffusion.
\newblock In \emph{NeurIPS}, 2023.
\newblock URL \url{https://arxiv.org/abs/2303.10834}.

\bibitem[Wu et~al.(2023{\natexlab{a}})Wu, Hu, Lu, Gilitschenski, and Garg]{wu2023slotdiffusion}
Ziyi Wu, Jingyu Hu, Wuyue Lu, Igor Gilitschenski, and Animesh Garg.
\newblock Slotdiffusion: Object-centric generative modeling with diffusion models.
\newblock In \emph{NeurIPS}, 2023{\natexlab{a}}.
\newblock URL \url{https://arxiv.org/abs/2305.11281}.

\bibitem[Bao et~al.(2023)Bao, Tokmakov, Wang, Gaidon, and Hebert]{bao2023motionguided}
Zhipeng Bao, Pavel Tokmakov, Yu-Xiong Wang, Adrien Gaidon, and Martial Hebert.
\newblock Object discovery from motion-guided tokens.
\newblock In \emph{CVPR}, 2023.
\newblock URL \url{https://arxiv.org/abs/2303.15555}.

\bibitem[Assran et~al.(2022)Assran, Caron, Misra, Bojanowski, Bordes, Vincent, Joulin, Rabbat, and Ballas]{Assran2022MSN}
Mahmoud Assran, Mathilde Caron, Ishan Misra, Piotr Bojanowski, Florian Bordes, Pascal Vincent, Armand Joulin, Michael~G. Rabbat, and Nicolas Ballas.
\newblock {Masked Siamese Networks for Label-Efficient Learning}.
\newblock In \emph{ECCV}, 2022.
\newblock URL \url{https://arxiv.org/abs/2204.07141}.

\bibitem[Chen et~al.(2021)Chen, Xie, and He]{Chen2021MoCov3}
Xinlei Chen, Saining Xie, and Kaiming He.
\newblock {An Empirical Study of Training Self-Supervised Vision Transformers}.
\newblock \emph{ICCV}, 2021.
\newblock URL \url{https://arxiv.org/abs/2104.02057}.

\bibitem[Qian et~al.(2023)Qian, Ding, Liu, and Lin]{qian2023semantics}
Rui Qian, Shuangrui Ding, Xian Liu, and Dahua Lin.
\newblock Semantics meets temporal correspondence: Self-supervised object-centric learning in videos.
\newblock In \emph{ICCV}, 2023.
\newblock URL \url{https://arxiv.org/abs/2308.09951}.

\bibitem[Aydemir et~al.(2023)Aydemir, Xie, and G{\"u}ney]{aydemir2023self}
G{\"o}rkay Aydemir, Weidi Xie, and Fatma G{\"u}ney.
\newblock Self-supervised object-centric learning for videos.
\newblock In \emph{NeurIPS}, 2023.
\newblock URL \url{https://arxiv.org/abs/2310.06907}.

\bibitem[Dosovitskiy et~al.(2021)Dosovitskiy, Beyer, Kolesnikov, Weissenborn, Zhai, Unterthiner, Dehghani, Minderer, Heigold, Gelly, Uszkoreit, and Houlsby]{Dosovitskiy2021ViT}
Alexey Dosovitskiy, Lucas Beyer, Alexander Kolesnikov, Dirk Weissenborn, Xiaohua Zhai, Thomas Unterthiner, Mostafa Dehghani, Matthias Minderer, Georg Heigold, Sylvain Gelly, Jakob Uszkoreit, and Neil Houlsby.
\newblock {An Image is Worth 16x16 Words: Transformers for Image Recognition at Scale}.
\newblock In \emph{ICLR}, 2021.
\newblock URL \url{https://openreview.net/forum?id=YicbFdNTTy}.

\bibitem[Tangemann et~al.(2023)Tangemann, Schneider, von Kügelgen, Locatello, Gehler, Brox, Kümmerer, Bethge, and Schölkopf]{Tangemann2023commonfate}
Matthias Tangemann, Steffen Schneider, Julius von Kügelgen, Francesco Locatello, Peter Gehler, Thomas Brox, Matthias Kümmerer, Matthias Bethge, and Bernhard Schölkopf.
\newblock Unsupervised object learning via common fate.
\newblock In \emph{CLeaR}, 2023.
\newblock URL \url{https://arxiv.org/abs/2110.06562}.

\bibitem[Jabri et~al.(2020)Jabri, Owens, and Efros]{jabri2020crw}
Allan Jabri, Andrew Owens, and Alexei Efros.
\newblock Space-time correspondence as a contrastive random walk.
\newblock In \emph{NeurIPS}, 2020.
\newblock URL \url{https://arxiv.org/abs/2006.14613}.

\bibitem[Yang et~al.(2019)Yang, Fan, and Xu]{vis2019}
Linjie Yang, Yuchen Fan, and Ning Xu.
\newblock Video instance segmentation.
\newblock In \emph{ICCV}, 2019.
\newblock URL \url{https://arxiv.org/abs/1905.04804}.

\bibitem[Pont-Tuset et~al.(2017{\natexlab{a}})Pont-Tuset, Perazzi, Caelles, Arbel\'aez, Sorkine-Hornung, and {Van Gool}]{davis2017}
Jordi Pont-Tuset, Federico Perazzi, Sergi Caelles, Pablo Arbel\'aez, Alexander Sorkine-Hornung, and Luc {Van Gool}.
\newblock The 2017 davis challenge on video object segmentation.
\newblock \emph{arXiv:1704.00675}, 2017{\natexlab{a}}.
\newblock URL \url{https://arxiv.org/abs/1704.00675}.

\bibitem[Lin et~al.(2014)Lin, Maire, Belongie, Hays, Perona, Ramanan, Doll{\'a}r, and Zitnick]{Lin2014COCO}
Tsung-Yi Lin, Michael Maire, Serge~J. Belongie, James Hays, Pietro Perona, Deva Ramanan, Piotr Doll{\'a}r, and C.~Lawrence Zitnick.
\newblock {Microsoft COCO: Common Objects in Context}.
\newblock In \emph{ECCV}, 2014.
\newblock URL \url{https://arxiv.org/abs/1405.0312}.

\bibitem[Pont-Tuset et~al.(2017{\natexlab{b}})Pont-Tuset, Arbeláez, Barron, Marques, and Malik]{Arbelaez2014MCG}
Jordi Pont-Tuset, Pablo Arbeláez, Jonathan~T. Barron, Ferran Marques, and Jitendra Malik.
\newblock {Multiscale Combinatorial Grouping for Image Segmentation and Object Proposal Generation}.
\newblock \emph{{IEEE} Transactions on Pattern Analysis and Machine Intelligence}, 39\penalty0 (1), 2017{\natexlab{b}}.
\newblock \doi{10.1109/TPAMI.2016.2537320}.
\newblock URL \url{https://ieeexplore.ieee.org/document/7423791}.

\bibitem[Rolfe(2017)]{rolfe2017discrete}
Jason~Tyler Rolfe.
\newblock Discrete variational autoencoders.
\newblock In \emph{ICLR}, 2017.
\newblock URL \url{https://openreview.net/forum?id=ryMxXPFex}.

\bibitem[Gumbsch et~al.(2021)Gumbsch, Butz, and Martius]{gumbsch2021sparsely}
Christian Gumbsch, Martin~V Butz, and Georg Martius.
\newblock Sparsely changing latent states for prediction and planning in partially observable domains.
\newblock In \emph{NeurIPS}, 2021.
\newblock URL \url{https://arxiv.org/abs/2110.15949}.

\bibitem[Wu et~al.(2023{\natexlab{b}})Wu, Dvornik, Greff, Kipf, and Garg]{wu2023slotformer}
Ziyi Wu, Nikita Dvornik, Klaus Greff, Thomas Kipf, and Animesh Garg.
\newblock Slotformer: Unsupervised visual dynamics simulation with object-centric models.
\newblock In \emph{ICLR}, 2023{\natexlab{b}}.
\newblock URL \url{https://openreview.net/forum?id=TFbwV6I0VLg}.

\bibitem[Zadaianchuk et~al.(2020)Zadaianchuk, Seitzer, and Martius]{zadaianchuk2020self}
Andrii Zadaianchuk, Maximilian Seitzer, and Georg Martius.
\newblock {Self-supervised Visual Reinforcement Learning with Object-centric Representations}.
\newblock In \emph{ICLR}, 2020.
\newblock URL \url{https://openreview.net/forum?id=xppLmXCbOw1}.

\bibitem[Mambelli et~al.(2022)Mambelli, Tr{\"a}uble, Bauer, Sch{\"o}lkopf, and Locatello]{mambelli2022compositional}
Davide Mambelli, Frederik Tr{\"a}uble, Stefan Bauer, Bernhard Sch{\"o}lkopf, and Francesco Locatello.
\newblock {Compositional Multi-object Reinforcement Learning with Linear Relation Networks}.
\newblock In \emph{ICLR Workshop on the Elements of Reasoning: Objects, Structure and Causality}, 2022.
\newblock URL \url{https://openreview.net/forum?id=HFUxPr_I5ec}.

\bibitem[Feng and Magliacane(2023)]{feng2023learning}
Fan Feng and Sara Magliacane.
\newblock Learning dynamic attribute-factored world models for efficient multi-object reinforcement learning.
\newblock In \emph{NeurIPS}, 2023.
\newblock URL \url{https://arxiv.org/abs/2307.09205}.

\bibitem[Salehi et~al.(2023)Salehi, Gavves, Snoek, and Asano]{Salehi_2023_ICCV}
Mohammadreza Salehi, Efstratios Gavves, Cees~G.M. Snoek, and Yuki~M. Asano.
\newblock Time does tell: Self-supervised time-tuning of dense image representations.
\newblock In \emph{ICCV}, 2023.
\newblock URL \url{https://arxiv.org/abs/2308.11796}.

\bibitem[Xie et~al.(2022)Xie, Xie, and Zisserman]{xie2022segmenting}
Junyu Xie, Weidi Xie, and Andrew Zisserman.
\newblock Segmenting moving objects via an object-centric layered representation.
\newblock In \emph{NeurIPS}, 2022.
\newblock URL \url{https://arxiv.org/abs/2207.02206}.

\bibitem[Traub et~al.(2023{\natexlab{b}})Traub, Becker, Otte, and Butz]{traub2023looping}
Manuel Traub, Frederic Becker, Sebastian Otte, and Martin~V Butz.
\newblock Looping loci: Developing object permanence from videos.
\newblock 2023{\natexlab{b}}.
\newblock URL \url{https://arxiv.org/abs/2310.10372}.

\bibitem[Oquab et~al.(2023)Oquab, Darcet, Moutakanni, Vo, Szafraniec, Khalidov, Fernandez, Haziza, Massa, El-Nouby, Howes, Huang, Xu, Sharma, Li, Galuba, Rabbat, Assran, Ballas, Synnaeve, Misra, Jegou, Mairal, Labatut, Joulin, and Bojanowski]{oquab2023dinov2}
Maxime Oquab, Timothée Darcet, Theo Moutakanni, Huy~V. Vo, Marc Szafraniec, Vasil Khalidov, Pierre Fernandez, Daniel Haziza, Francisco Massa, Alaaeldin El-Nouby, Russell Howes, Po-Yao Huang, Hu~Xu, Vasu Sharma, Shang-Wen Li, Wojciech Galuba, Mike Rabbat, Mido Assran, Nicolas Ballas, Gabriel Synnaeve, Ishan Misra, Herve Jegou, Julien Mairal, Patrick Labatut, Armand Joulin, and Piotr Bojanowski.
\newblock {D}{I}{N}{O}v2: Learning robust visual features without supervision, 2023.
\newblock URL \url{https://arxiv.org/abs/2304.07193}.

\end{thebibliography}

\appendix

\counterwithin{figure}{section}
\counterwithin{table}{section}

\begin{center}
\LARGE 
Supplementary Material for \\Object-Centric Learning for Real-World Videos by Predicting Temporal Feature Similarities
\end{center}

\startcontents[sections]
\printcontents[sections]{l}{1}{\setcounter{tocdepth}{2}}

\newpage

\section{Comparison with Additional Baselines}

\subsection{Comparison with Image-Based Object-Centric Methods}
\label{app:image_comparison}
\setcounter{figure}{0}

In this section, we evaluate how effective our model is for unsupervised \emph{image} segmentation from videos.
In addition to reporting results for the video-based object-centric methods SAVi and STEVE, we compare with several recent image-based object-centric learning methods. SLATE~\citep{Singh2022SLATE} is an image-based object-centric model that trains a discrete VAE~\citep{rolfe2017discrete} as a dense feature extractor and uses a Transformer decoder conditioned on slots to reconstruct discrete representations of VAE features. 
LSD~\citep{jiang2023object} replaces the Transformer decoder with a latent diffusion model conditioned on the object slots. 
Next, DINOSAUR~\citep{seitzer2023bridging} incorporates dense DINO features as targets and reconstructs the features itself. 
We report the \textit{Image FG-ARI} and \textit{Image mBO} metrics. 
They measure how well the predicted segmentation matches the ground-truth segmentation of a given single image~(frame), thus consistency over the video is not taken into account.

The results for MOVi datasets are presented in \tab{tab:combined_image}. 
\method surpasses both previous image- and video-based methods, showing the benefits of using motion information in combination with semantically coherent self-supervised features. 
Interestingly, \method performs well on both Image FG-ARI and Image mBO metrics, whereas DINOSAUR and LSD seem to improve either in quality of split (measured in Image FG-ARI) or in the sharpness of masks (measured in Image mBO) at the cost of performing worse on the second metric. 
We also note that LSD results are from larger resolution of MOVi images ($256 \times 256$). 
We expect that our method can additionally improve if we also use larger resolution videos; however, to be comparable with other baselines, we use $128 \times 128$ resolution in this work.

\begin{wrapfigure}[8]{r}{0.5\textwidth}
    \centering
    \vspace{-1.5em}
    \includegraphics[width=0.48\textwidth]{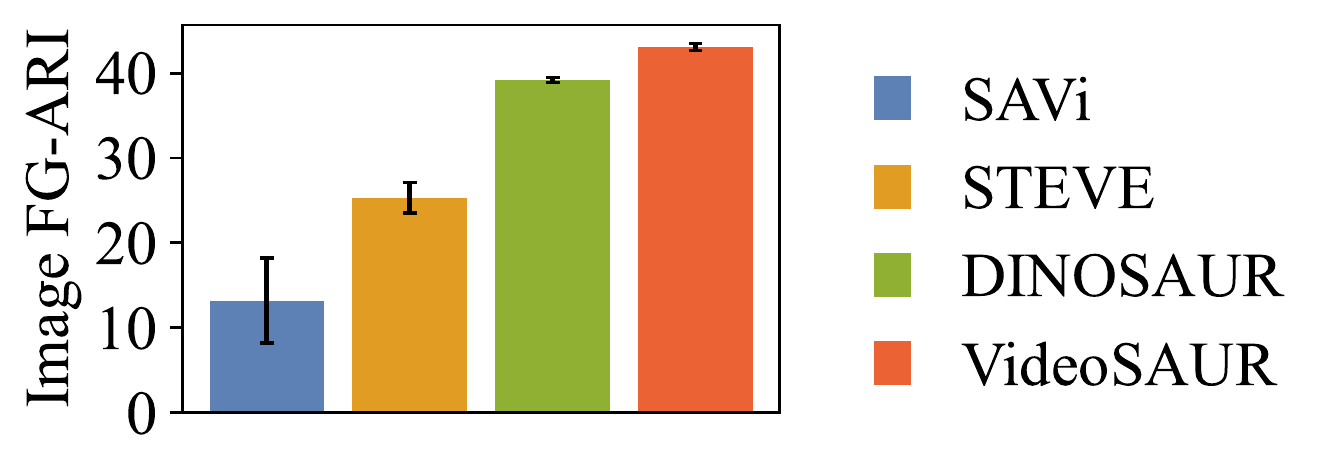}
    \caption{Image-based comparison on YouTube-VIS~(mean $\pm$ standard dev., 3 seeds).}
    \label{fig:image_based_ytvis}
\end{wrapfigure}
In addition, we also compare \method with DINOSAUR and other video-based baselines on the more challenging YouTube-VIS dataset~(see \fig{fig:image_based_ytvis}). 
\method outperforms DINOSAUR ($+4$ FG-ARI) and also surpasses video-based STEVE and SAVi methods. 
This underscores the benefit of our temporal similarity loss over mere feature reconstruction for challenging real-world datasets.

\begin{table}[h]
\centering
\caption{Comparison with state-of-the-art methods on the MOVi-C, MOVi-E image datasets. Both metrics are computed for individual frames. The results for SLATE and DINOSAUR are from \citet{seitzer2023bridging}, while LSD results are from ~\citet{jiang2023object}. We report mean $\pm$ standard dev. over 5 runs for our model.}
\label{tab:combined_image}
\setlength{\tabcolsep}{5pt}
\begin{tabular}{@{}lrrrrrr@{}}\toprule
&\multicolumn{2}{c}{MOVi-C} && \multicolumn{2}{c}{MOVi-E}  \\\cmidrule{2-3}\cmidrule{5-6}
&Image FG-ARI & Image mBO &&Image FG-ARI & Image mBO   \\\midrule
Block Pattern &42.7 &19.5 &&41.9 &20.4  \\
SAVi ~\citep{kipf2022conditional} &41.8	& 25.9&& 50.3 &	20.3 \\
STEVE~\citep{Singh2022STEVE} &51.9 &41.6  && 59.5	& 34.4   \\
SLATE~\citep{Singh2022SLATE} &43.6 &26.5 && 44.4 &23.6  \\
LSD~\citep{jiang2023object}  &50.5 &46.3 && 53.4 &39.6   \\
SlotDiffusion~\citep{wu2023slotdiffusion} & -- & -- && 60.0 & 30.2 \\
DINOSAUR~\citep{seitzer2023bridging} &68.6 &39.1 && 65.1 &35.5  \\
\method & \textbf{75.5 ± 0.9 }&	\textbf{46.0 ± 0.6} && \textbf{78.4 ± 0.7} &\textbf{ 41.2 ± 0.4} \\
\bottomrule
\end{tabular}
\end{table}

\subsection{Comparison with Concurrent Work on Real-World Videos}
\label{app:concurrent_work_comparison}

In concurrent work, two more slot attention-based methods were proposed that learn object-centric representations on real-world videos: SMTC~\cite{qian2023semantics} and SOLV~\citep{aydemir2023self}.
SMTC learns to extracts objects from videos by enforcing semantic and instance consistency over time using a student-teacher approach.
SOLV extracts per-frame slots using invariant slot attention~\citep{Biza2023invariant}, applies a temporal consistency module and merges slots using agglomerative clustering; the model is trained using DINOSAUR-style feature reconstruction on masked out intermediate frames.
In this section, we compare to them using the respective evaluation protocols in their papers.

First, we compare to SMTC in \tab{tab:smtc_comparison} on the DAVIS-2017-Unsupervised dataset~\citep{davis2017}.
Specifically, we assess our method’s transfer performance when trained on YT-VIS 201, and follow the evaluation procedure outlined in~\citep{davis2017} for matching ground truth masks to predictions using mean $\mathcal{J}$ \& $\mathcal{F}$.
We report the Jaccard index $\mathcal{J}$ (equivalent to IoU), the boundary F-score $\mathcal{F}$ and their average as $\mathcal{J}$ \& $\mathcal{F}$.
While the contours of \method's segmentation masks (as measured by $\mathcal{F}$) are not as accurate due to processing images and predicting masks at a lower patch resolution ($37 \times 37$ for \method trained with DINOv2 B14 features on original DINOv2 resolution), the Jaccard index $\mathcal{J}$ is comparable to SMTC.

Second, we compare to SOLV in \tab{tab:solv_comparison} on the YT-VIS 2019 dataset.
We also list the performance of OCLR~\citep{xie2022segmenting}, which uses synthetic data with ground-truth optical flow.
As SOLV, we report the mIoU metric matched over the entire video.
The results for \method are obtained using random $200$ videos from the YT-VIS 2019 train split \footnote{We use part of the train split, as validation labels are not released. Exact indexes are available at \texttt{validation} split of the Tensorflow version of the YT-VIS 2019 dataset: \url{https://www.tensorflow.org/datasets/catalog/youtube_vis\#youtube_visonly_frames_with_labels_train_split}}.
Our results surpass OCLR, showing \method’s effectiveness in extracting motion information directly from video data using dense self-supervised features. 
Additionally, our performance matches that of SOLV without using agglomerative clustering, while SOLV with slot merging outperforms \method ($+5$ mIoU).
This highlights the importance of correctly determining the number of slots. 
While the main concern in this paper is to integrate motion information from video, we see determining the number of slots as an important orthogonal direction.
Thus, combining our method a solution such as the one from~\citet{aydemir2023self} is an interesting direction for future work. 

\begin{table}
  \centering
  \setlength{\tabcolsep}{5pt}
  \begin{minipage}[t]{0.48\linewidth}
    \centering
    \caption{Comparison of \method with DINOv2 ViT B/14 features to SMTC~\cite{qian2023semantics} on the DAVIS-2017-Unsupervised validation set. 
    The results for SMTC are from~\citet{qian2023semantics}.} \label{tab:smtc_comparison}
    \begin{tabular}{@{}lccc@{}}
      \toprule
      \multicolumn{1}{c}{Method}  & $\mathcal{J}$ & $\mathcal{F}$ & $\mathcal{J}$ \& $\mathcal{F}$ \\
      \midrule
      SMTC & 36.4 & \textbf{44.6}& \textbf{40.5} \\
      \method & \textbf{36.8} & 21.1 & 29.0 \\
      \bottomrule
    \end{tabular}
  \end{minipage}%
  \hfill
  \begin{minipage}[t]{0.48\linewidth}
    \centering
    \caption{Comparison of \method with DINOv2 ViT B/14 features to OCLR~\citep{xie2022segmenting} and SOLV~\citep{aydemir2023self} on YT-VIS 2019. 
    The results for OCLR and SOLV are from~\citet{aydemir2023self}.} \label{tab:solv_comparison}
    \begin{tabular}{@{}lc@{}}
      \toprule
      \multicolumn{1}{c}{Method}  & mIoU \\
      \midrule
      OCLR & 32.5 \\
      SOLV (w/o slot merging) & 39.9 \\
      SOLV (w/ slot merging) & \textbf{45.3} \\
      \method & 40.3 \\
      \bottomrule
    \end{tabular}
  \end{minipage}
\end{table}

\section{Additional Experiments}
\label{app:more_exeriments}
\setcounter{figure}{0}

\subsection{Long-Term Video Consistency}
\label{app:video_length}
\begin{figure}
    \centering
    \includegraphics[width=0.6\textwidth]{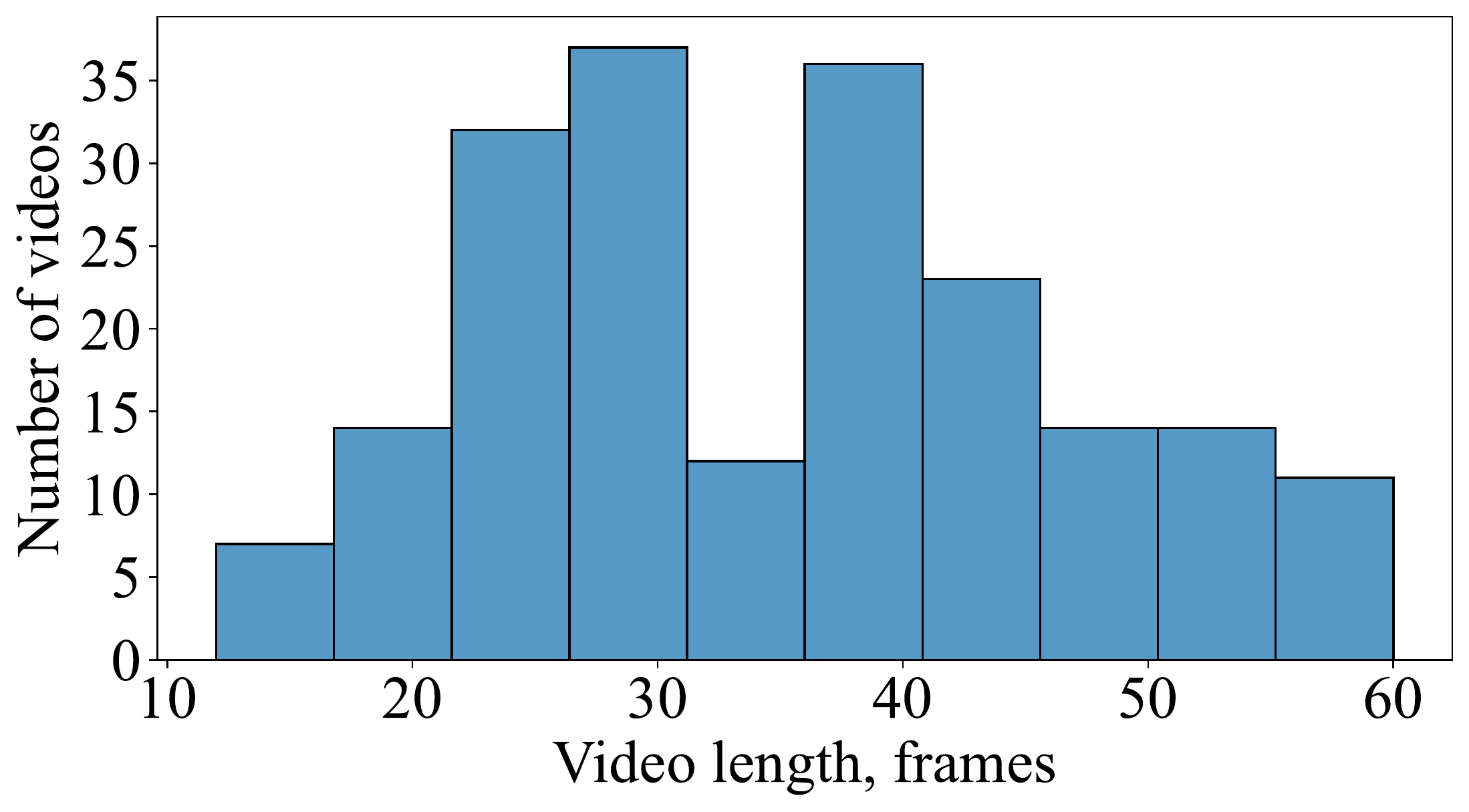}
    \caption{Histogram of video lengths on the YT-VIS 2021 validation dataset.}
    \label{fig:ytvis2021_train_video_len}
\end{figure}

Beyond the initial examination of how our \method performs on the relatively brief 6-frame video segments from YouTube-VIS 2021, we extend our evaluation to also assess its effectiveness on more substantial, longer video segments. 
In \tab{tab:video_length}, we show the performance for 12-frame and full YT-VIS video segments (see \fig{fig:ytvis2021_train_video_len} for the distribution of video lengths). 
Although the performance of \method on extended video segments predictably decreases in terms of FG-ARI (as we do not have any memory module to handle object occlusions and reidentification~\citep{traub2023looping}), the observed difference between \method and the baseline models is significant. 
This suggests that \method maintains its efficacy in tracking the primary objects in videos across longer time intervals. 

Additionally, we investigate if \method benefits from using DINOv2 features~\citep{oquab2023dinov2} that are obtained by training DINO on the larger dataset and fine-tuning the representation on larger resolution ($518\times 518$). We show that \method performance benefits from using such features as a backbone, especially in terms of mask quality ($+6$ mBO points). Using those features with the original resolution \method reaching $29.7$ mBO on full-length YT-VIS 2021 videos.  In addition to this quantitative evaluation, we visualize \method predictions on long YT-VIS videos (longer than 30-frames) in \Fig{fig:ytvis2021_examples_long} and \Fig{fig:ytvis2021_examples_long_dinov2}.

\begin{table}
\centering
\caption{Performance of \method on the YT-VIS 2021 dataset, varying the length of the video segment (mean $\pm$ standard dev., 5 seeds for \method with DINO features and STEVE. \method with DINOv2 features are one seed only due to computational limitations).}
\label{tab:video_length}
\small
\setlength{\tabcolsep}{2.7pt}
\begin{tabular}{@{}lrrrrrrrr@{}}\toprule
&\multicolumn{2}{c}{6 frames} && \multicolumn{2}{c}{12 frames} && \multicolumn{2}{c}{Full Videos} \\\cmidrule{2-3}\cmidrule{5-6}\cmidrule{8-9}
&FG-ARI & mBO &&FG-ARI & mBO &&FG-ARI & mBO \\\midrule
Block Pattern &24.0	&14.9 &&20.3	&14.2&& 15.1 & 13.1 \\
STEVE~\citep{Singh2022STEVE} & 20.0 ± 1.5	& 20.9 ± 0.5 && 18.0 ± 1.4	& 21.5 ± 0.5 && 15.0 ± 0.7	& 19.1 ± 0.4 \\
\method with DINO  B/16& 39.5 ± 0.6 & 29.1 ± 0.4 && 35.8 ± 0.3	& 29.4 ± 0.3 && 28.9 ± 0.4 & 26.3 ± 0.2 \\
\method with DINOv2 B/14~\citep{oquab2023dinov2} & 39.7 & 35.6 && 38.7	& 34.5 && 31.2 & 29.7\\
\bottomrule
\end{tabular}
\end{table}

\subsection{Optical Flow as Self-Supervised Target}
\label{app:optical_flow}
The choice of the self-supervised target plays an important role in creating suitable inductive biases for object discovery and scene decomposition. 
As such, understanding the properties of the prediction target leading to an effective scene decomposition is crucial. 
The temporal feature similarity prediction proposed in this work combines two different inductive biases: high-level semantic information and motion information. 
To elucidate the significance of both types of bias for scene decomposition, we assess the performance of VideoSAUR with prediction targets that consist only of one of those biases.

In \tab{tab:loss_sim_alation_movic} and \tab{tab:loss_sim_alation_ytvis} in the main text, we compare predicting temporal similarities with predicting self-supervised features of the current frame.
Such features only contain semantic information, but no information about motion.
Depending on the dataset, including temporal information brings a small (YT-VIS) or large benefit (MOVi-C).

Subsequently, we study if motion cues alone (without the semantic information) are enough for successful scene decomposition.  
In particular, we compare self-supervised temporal similarity targets with (ground truth) optical flow targets (only motion information) on the MOVi-C and MOVi-E datasets. 
To this end, we train \method by predicting optical flow targets, using a spatial broadcast decoder similar to SAVi~\citep{kipf2022conditional} instead of the mixer decoder.
All other components of \method stay unchanged.
The optical flow map is predicted at full image resolution ($128 \times 128$).

The results are presented in \tab{tab:movi_comparison_optical_flow}. 
We train VideoSAUR with GT optical flow for the best potential performance from optical flow alone. 
Yet, even on the MOVi-C datasets favoring optical flow (no camera motion, no static objects), VideoSAUR with temporal feature similarities significantly outperforme optical flow (+10 FG-ARI). 
This disparity is even greater on the MOVi-E dataset, highlighting VideoSAUR's resilience to camera movements and static objects. 
Together, these results demonstrate that our temporal feature similarity targets, despite not requiring signals such as optical flow (which would need estimation in real-world scenarios like the YouTube-VIS dataset), excel over mere optical flow targets.
We attribute this to the enriched semantic bias inherent to the self-supervised feature similarities.\looseness=-1

\begin{table}
  \centering
  \caption{Comparing \method predicting temporal similarities to predicting ground truth optical flow on the MOVi-C and MOVi-E datasets. We report Video FG-ARI of a version of \method with optical flow (both backward and forward) as well as the original VideoSAUR with temporal features similarity.}
  \label{tab:movi_comparison_optical_flow}
  \setlength{\tabcolsep}{5pt}
  \begin{tabular}{@{}lcc@{}}
    \toprule
    \multicolumn{1}{c}{VideoSAUR}  & MOVi-C  & MOVi-E\\
    \midrule
    w/ GT Optical Flow (backward) & 48.1 & 28.9 \\
    w/ GT Optical Flow (forward)  & 48.9 & 30.1 \\
    w/ Temporal Similarities  & 60.7 & 73.9 \\
    \bottomrule
  \end{tabular}
\end{table}

\subsection{Comprehensive Ablation Study on the MOVi-E Dataset}
\label{app:movi_e}

In this section, we present the results of a comprehensive ablation study conducted on the MOVi-E dataset. The purpose of this study is to investigate the impact of two key factors: decoder choice (MLP vs. Mixer) and loss function selection. Our goal is to gain a deeper understanding of how these choices affect the performance of our method. The results are summarized in \tab{tab:model_comparison_movi_e}.

\begin{table}
    \centering
    \caption{Extended ablation of VideoSAUR components on MOVi-E. We compare VideoSAUR model with different choices of the decoder (Mixer vs MLP used by DINOSAUR) and loss types 
    (temporal similarity loss vs feature reconstruction).
    }
    \label{tab:model_comparison_movi_e}
    \setlength{\tabcolsep}{3pt}
    \begin{tabular}{@{}lllrr@{}}
    \toprule
    Decoder & Loss  Type& & FG-ARI & mBO \\
    \midrule
    MLP     & Feature Reconstruction & & 68.6 & 27.6 \\
    MLP     & Temp. Feat. Sim. Prediction & & 74.5 & 28.8 \\
    Mixer   & Feature Reconstruction & & 62.3 & 20.6 \\
    Mixer   & Temp. Feat. Sim.  Prediction& & 74.1 & 34.1 \\
    \bottomrule
    \end{tabular}
\end{table}

The similarity loss is beneficial for both decoders, pushing the FG-ARI to approximately $74$, as compared to $69$ when using the feature reconstruction loss. Notably, the Mixer decoder significantly enhances the clarity of the object mask, with an improvement of $+5$ mBO. When combined, the similarity loss and Mixer consistently outshine the MLP decoder equipped with the feature reconstruction loss. These insights provide valuable auxiliary information to our main paper ablations (see \Tab{tab:loss_sim_alation_movic} and \Tab{tab:loss_sim_alation_ytvis}), painting a more detailed picture of \method's components and their respective performances.

\subsection{Stability of Mixer Decoder}
\label{app:mixer_stability}

\begin{table}
  \centering
    \begin{minipage}[b]{0.48\linewidth}
    \centering
    \captionof{table}{Mixer decoder with smaller DINO features (S/16) on YT-VIS 2021~(mean $\pm$ standard dev., 3 seeds).}\label{tab:mixer_stability}
    \small
    \setlength{\tabcolsep}{4.7pt}
    \begin{tabular}{@{}cc|rrr@{}}
      \multicolumn{2}{c}{Loss type} & \multicolumn{2}{c}{Metric} \\
      \toprule
      Feat. Rec. & Temp. Sim. & FG-ARI & mBO \\
      \midrule
      \checkmark & &14.9 ± 12.0	& 12.9 ± 5.9 \\
      \checkmark & \checkmark &37.0 ± 3.5& 29.1 ± 0.6\\
      \bottomrule
    \end{tabular}
  \end{minipage}%
  \hfill
  \begin{minipage}[b]{0.48\linewidth}
    \centering
    \captionof{table}{Loss ablation study on COCO dataset. Metrics are image-based ARI and mBO~(mean, 3 seeds).}\label{tab:loss_sim_alation_coco}
    \small
    \begin{tabular}{@{}cc|rrr@{}}
      \multicolumn{2}{c}{Loss type} & \multicolumn{2}{c}{Metric} \\
      \toprule
      Feat. Rec. &Self-Sim. & FG-ARI & mBO$^{i}$ \\
      \midrule
      \checkmark & &34.8 &23.9 \\
      &\checkmark &28.5 &25.6 \\
      \checkmark&\checkmark &38.0 &25.9 \\
      \bottomrule
    \end{tabular}
  \end{minipage}%
\end{table}

As mentioned in \sect{subsec:method-slotmixer} of the main text, we found that the mixer decoder sometimes exhibits training instabilities.
For instance, \tab{tab:mixer_stability} shows that there is high variance over random seeds when training purely with feature reconstruction, \ie some training runs fail to discover an object grouping.
These instabilities manifest in slot masks that follow a Voronoi-like decomposition of the image.
When adding the temporal similarity loss, the instabilities disappear.

We hypothesize this is because the mixer decoder has increased flexibility in how to model the image with slots (compared to the conventional mixture-based decoder), and thus more failure modes (non-object groupings) the model can ``fall into'' during training.
Increasing the difficulty of the task by adding the temporal similarity loss makes these failure modes less viable: by putting more pressure on the slot bottleneck to encode information, object-based slot groupings are more efficient representations than alternative groupings.

However, we found that the mixer decoder with feature reconstruction does not show instabilities in \emph{all} settings. 
For example, training with DINO ViT Base/8 features on MOVi-C or DINO Base/16 features on YouTube-VIS 2021 (\fig{tab:loss_sim_alation_movic} and \fig{tab:loss_sim_alation_ytvis} in the main text) is relatively stable.
We attribute this to the increased task difficulty when predicting ViT ``Base'' features instead of ViT ``Small'' features (as in \tab{tab:mixer_stability}). 
Once more~\citep{kipf2022conditional,Singh2022SLATE,yang2022promising,elsayed2022savi++, seitzer2023bridging}, these findings demonstrate the central lesson of unsupervised object discovery: to be successful, the model needs to have sufficient inductive biases, whether they stem from the dataset, the decoder, the grouping module, or the training task.

\subsection{Image-Based Feature Similarity on COCO}
\label{app:loss_sim_alation_coco}

In this section, we show that our proposed similarity loss can also be useful for image-based datasets, and thus is not restricted to just the video setting.
Note that by setting the time-shift $k$ to $0$, the temporal similarities turn into a \emph{self-similarities}, that is, the target similarities are computed by comparing features from the same image.
The resulting similarity maps highlight semantically and spatially related patches and thus could be useful targets to discover objects.

To test this, we train the DINOSAUR method with ViT S/16~\citep{seitzer2023bridging} with the self-similarity loss on the real-world COCO dataset (see \tab{tab:loss_sim_alation_coco}). 
Similar to the results from the time-shift analysis (see $k=0$ in \fig{fig:time_shift}), we find that using the self-similarity loss alone does not seem to carry enough signal to train the model and leads to degraded performance. 
However, combining the self-similarity loss with feature reconstruction shows significant improvement over using only feature reconstruction. 
Even though the targets contain the same information overall, different transformations of the original targets (e.g. their relative similarity) create different biases --- a combination of these \emph{different views into the targets} appears to be beneficial for object discovery.

\subsection{Sensitivity to Number of Slots During Evaluation}

\begin{figure}[hb]
    \centering
    \includegraphics[width=0.8\textwidth]{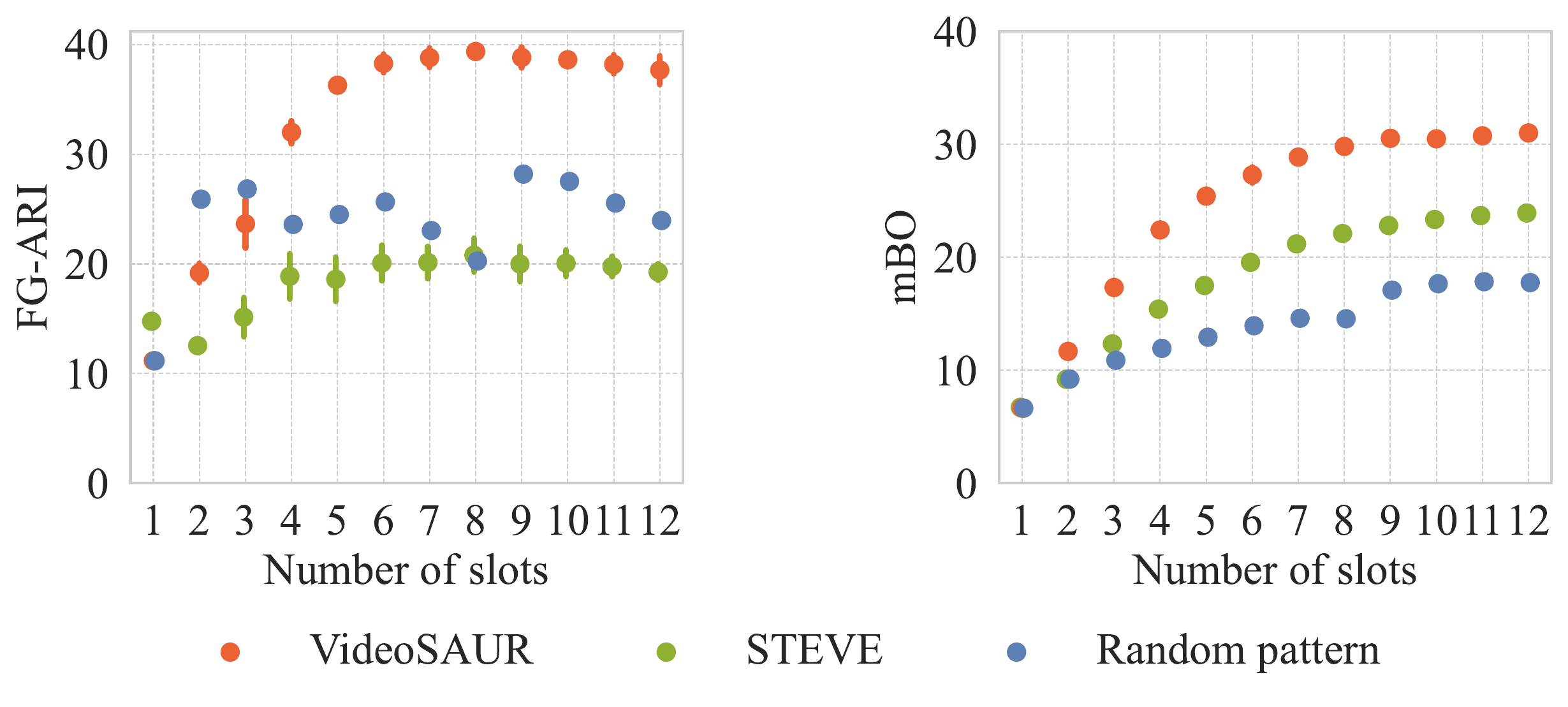}
    \caption{Changing the number of slots during evaluation on the YT-VIS 2021 dataset~(mean $\pm$ standard dev., 5 seeds).}
    \label{fig:n_slots_ytvis}
\end{figure}

One of the noteworthy properties of slot attention-based models with randomly sampled initialization\footnote{This is in contrast to the fixed learned initialization used in unconditioned SAVi~\citep{kipf2022conditional} and SAVi++\citep{elsayed2022savi++}.} is that the number of slots can be adjusted during inference. 
As we demonstrate in \fig{fig:time_shift}, this is helpful for successfully transferring to datasets that have a different average number of objects per video. For this purpose, it is important to examine how stable the model's performance is if it is used with a different number of slots than during training.
Thus, we evaluate our model (trained on YT-VIS 2021 with $k=7$ slots) using a varying number of slots (from 1 to 12) and present the results in \fig{fig:n_slots_ytvis}. 
We observe that while using fewer slots steadily deteriorates the performance, our method performs relatively well with a larger number of slots. 
This suggests that our method is relatively robust to the usage of a larger number of slots than needed. 
This property is useful for object discovery in images where the number of objects is unknown.

\subsection{Effect of ViT Architecture on Final Performance}
\begin{figure}[h]
\centering
\begin{minipage}{0.54\textwidth}
    \centering
    \includegraphics[width=\textwidth]
    {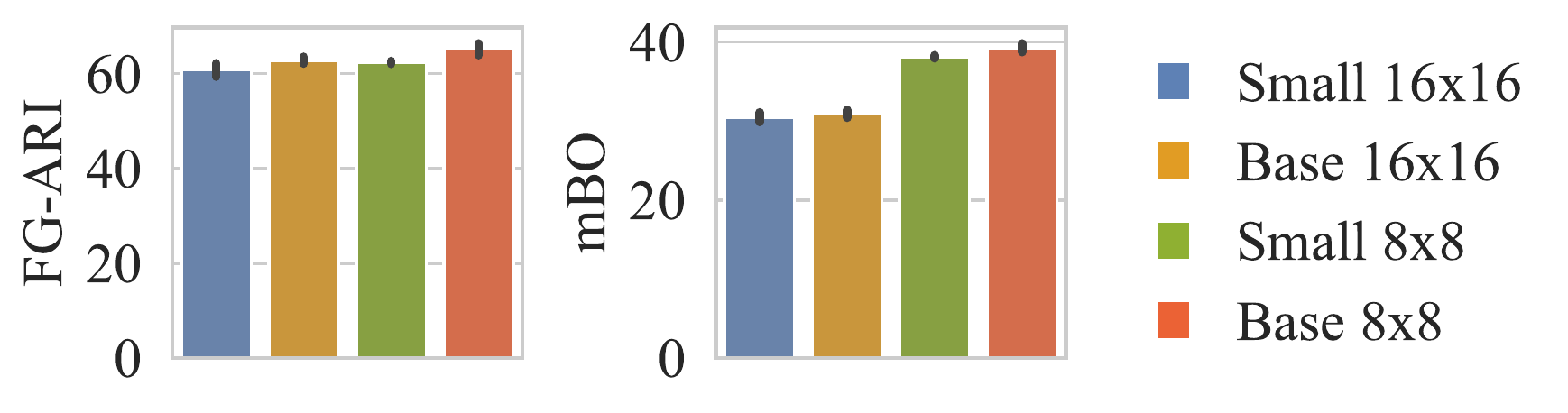}
    \captionof{figure}{Effect of ViT architecture choice and patch size on \method training on the MOVi-C~(mean $\pm$ standard dev., 3 seeds).}
    \label{fig:vit_type}
\end{minipage} \hfill
\begin{minipage}{0.45\textwidth}
\centering
\captionof{table}{ViT networks configuration.}
	\label{tab:vit_archs}
  \setlength{\tabcolsep}{2pt}
  \small
\begin{tabular}{@{}l c c c c c@{}}

\toprule
Model & Patch Size & Dim & Heads & Tokens & Params  \\
\midrule
Small &$16$  & 384 & 6 & 196 & 21M  \\
Small &$8$  & 384 & 6 & 784 & 21M  \\
Base &$16$ & 768 & 12 & 196 & 85M  \\
Base &$8$ & 768 & 12 & 784 & 85M  \\
\bottomrule
  \end{tabular}
\end{minipage}
\end{figure}

In addition to studying how the choice of self-superivsed method and ViT outputs affect the performance of \method (see \fig{fig:keys} and \fig{fig:features_choice} in the main paper), we also explore the effect of the \emph{scale of ViT architecture} on the performance.
To this end, we compare \method with DINO features of 2 different ViT sizes (Small and Base) and two different patch sizes ($16\times 16$ and $8\times8$ resolutions) on the MOVi-C dataset (see \tab{tab:vit_archs}). 
The results are presented in \fig{fig:vit_type}. 
We find that smaller patch size is important for sharper masks (measured by the mBO metric), while both larger architecture and smaller patch size are important for a better split of the scene to object masks (measured by the FG-ARI metric). %

\begin{figure}
    \centering
    \includegraphics[trim=0cm 0.3cm 0cm   0cm, clip, width=0.9\textwidth]{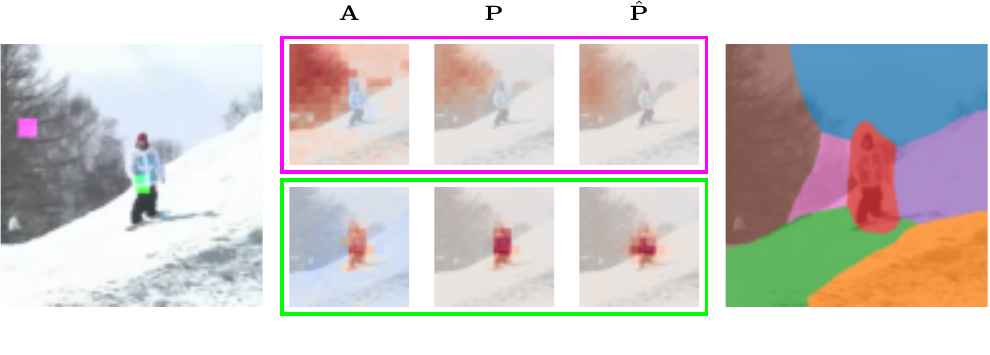}
    
        \includegraphics[trim=0cm 0.3cm 0cm   0cm, clip, width=.9\textwidth]{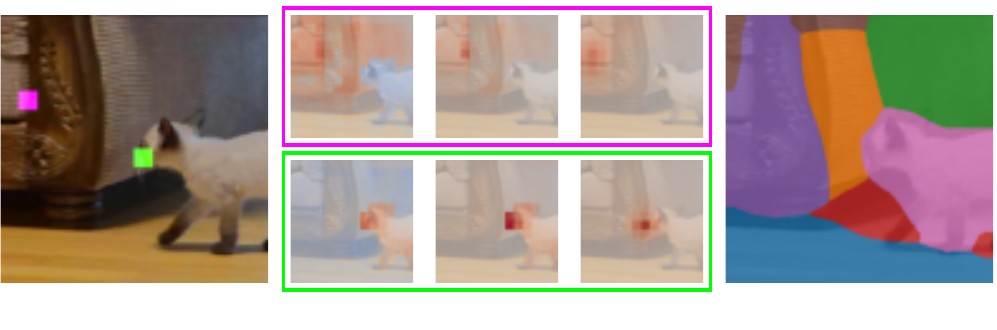}
            \includegraphics[width=0.9\textwidth]{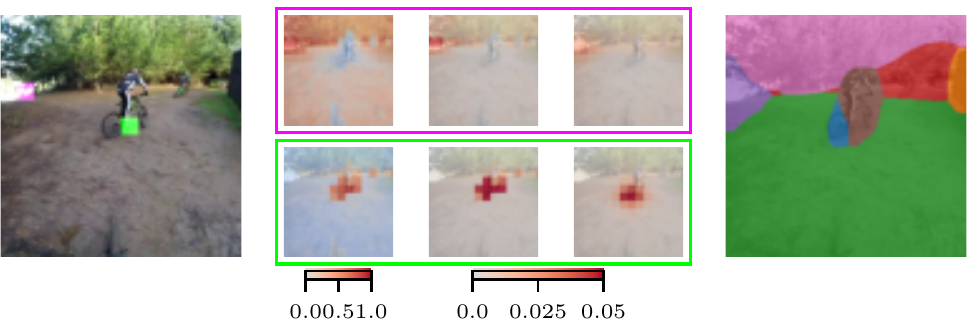}
      \caption{Additional visualization of affinity matrix $\mA$, transition probabilities $\mP$ and decoder predictions of transition probabilities $\hat{\mP}$ between patches (marked by purple and green) of the frame $\vx_t$ and patches of the next frame $\vx_{t+1}$ for YouTube-VIS 2021 validation videos. Red indicates maximum affinity/probability. }
    \label{fig:affinity_additional}
\end{figure}

\begin{figure}
  \centering
  \includegraphics[width=0.8\textwidth]{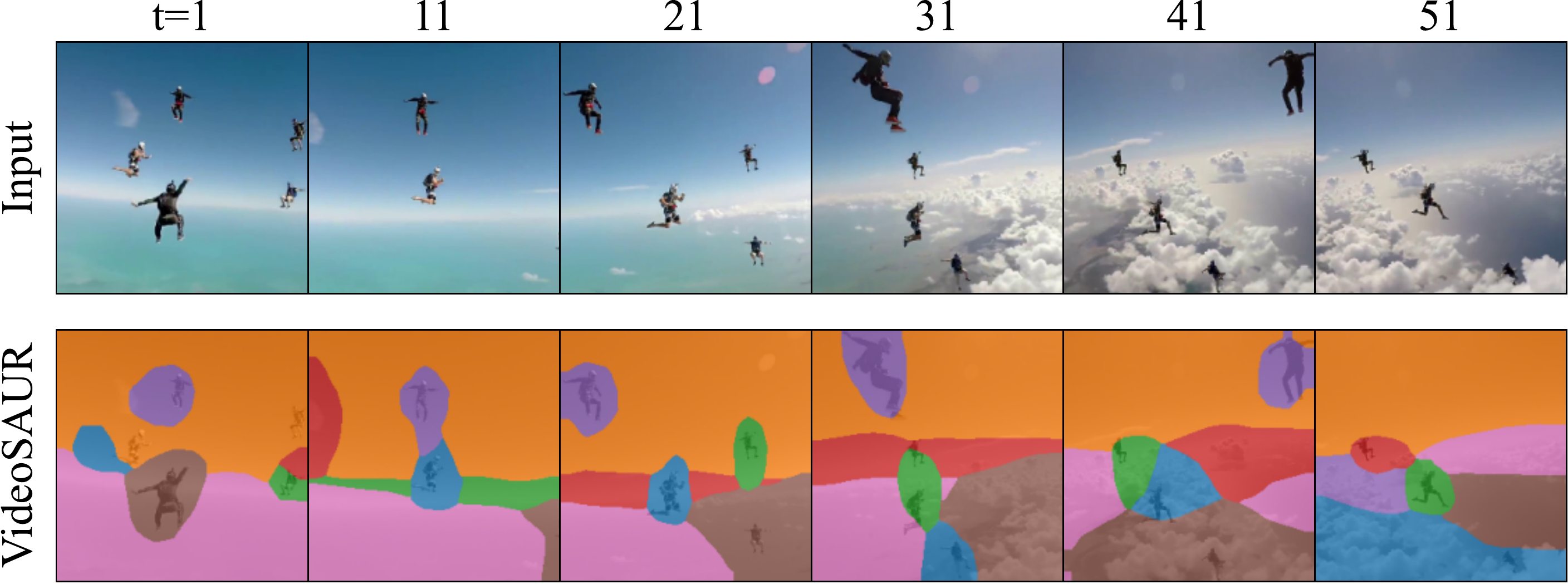}
    \caption{Failure case for long video prediction. Note that the original videos in YouTube-VIS are resampled from the original 30 fps to 6 fps, thus the original length of the video is 250 frames. 
    The slots are reassigned to the background, while small objects are not recognized.  
    }
    \label{fig:ytvis2021_fal_case}
\end{figure}

\section{Architectural Details and Hyperparameters}
Here we describe details about our model, its training and baselines that we use for comparison. 
We release our code at \url{https://github.com/martius-lab/videosaur}.
\setcounter{figure}{0}

\subsection{SlotMixer Decoder}
\label{app:mixer_decoder}

\begin{figure}[thb]
    \begin{minipage}{0.5\textwidth}
        \centering
        \includegraphics{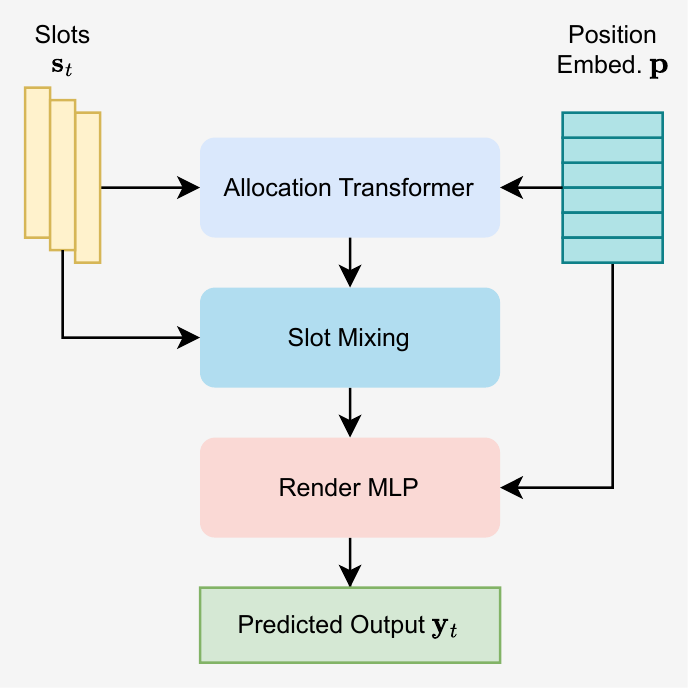}
    \end{minipage}
    \hfill
    \begin{minipage}{0.46\textwidth}
        \centering
        \begin{lstlisting}[language=Python]
pos_emb = Param(random_norm(n_pos, dim))

def mixer(slots):
    bs, n_slots, dim = slots.shape
    pos_emb = pos_emb.unsqueeze(0) \
                     .expand(bs, -1, -1)

    # Step 1: Allocation Transformer
    feats = xa_transf(q=pos_emb, 
                      kv=norm_kv(slots))

    # Step 2: Slot Mixing
    q = linear_q(norm_q(feats))
    k = linear_k(norm_k(slots))
    dots = einsum("bpd, bsd -> bps", 
                  q, k)
    dots *= q.shape[-1]**-0.5
    attn = softmax(dots, dim=-1)
    slot_mix = einsum("bps, bsd -> bpd", 
                      attn, slots)

    # Step 3: Rendering with MLP
    slot_mix += pos_emb
    outputs = mlp(slot_mix)

    return outputs\end{lstlisting}
    \end{minipage}
    \caption{SlotMixer decoder. Left: SlotMixer performs three steps for decoding: the allocation transformer assigns slots $\bm{s}_t$ to spatial positions $\bm{p}$, the slot mixing step creates a slot mix for each spatial position, and the render MLP decodes the slot mix to the final output $\bm{y}_t$. Right: PyTorch-like pseudocode for the SlotMixer decoder.}
    \label{fig:mixer}
\end{figure}

We describe the SlotMixer decoder, and how we adapted it for 2D decoding. 
We keep the original terminology from \citet{sajjadi2022object} for consistency.
SlotMixer performs three steps for decoding: the \emph{allocation step} assigns slots to spatial positions, the \emph{mixing step} creates a slot mix for each spatial position, and the \emph{render step} decodes the slot mix to the final output.
See \fig{fig:mixer} for an overview and pseudocode implementing the decoder.

\textbf{Allocation Step\quad}
This step takes as input the slots $\vs_t \in \mathbb{R}^{K \times M}$ and a learned positional embedding $\vp \in \mathbb{R}^{L \times M}$ and outputs a feature vector $\vf \in \mathbb{R}^{L \times M}$ using a cross-attention Transformer.
In particular, this Transformer iterates several cross-attention operations (and applies residual two-layer MLPs) using the position embeddings as queries to attend into the set of slots, where the position embeddings are residually updated using values from the slots.
Importantly, the position embeddings are processed independently of each other.
We utilize pre-normalization and also apply a layer norm to the slots before feeding them into the Transformer.
In contrast to \citet{sajjadi2022object}, we use a learned positional embedding initialized from a normal distribution instead of 3D encodings for the query rays, and do not apply a MLP to the positional embedding.

\textbf{Mixing Step\quad}
The mixing step is similar to a single-head attention step using the features $\vf$ as queries and the slots $\vs_t$ as keys, where the slots are averaged as the values to form the slot mix $\vm \in \mathbb{R}^{L \times M}$ that is used for decoding to the final output:
\begin{equation*}
\begin{array}{rlr}
\mathbf{q} &= \operatorname{norm}(\vf)\, U_q  & U_q \in \mathbb{R}^{M \times M}, \\
\mathbf{k} &= \operatorname{norm}(\vs)\, U_k  & U_k \in \mathbb{R}^{M \times M}, \\
\mA & =\operatorname{softmax}\left(\mathbf{q} \textbf{k}^{\top} / \sqrt{M}\right) & \mA \in \mathbb{R}^{L \times K}, \\
\vm & = \vs \mA & \vm \in \mathbb{R}^{L \times M}. \\
\end{array}    
\end{equation*}

\textbf{Render Step\quad}
The render step takes the slot mix $\vm \in \mathbb{R}^{L \times M}$, adds the positional embedding $\vp \in \mathbb{R}^{L \times M}$ and applies a MLP with ReLU activation independently to each position:
\begin{equation*}
\begin{array}{rl}
\vy &= \operatorname{MLP}(\vm + \vp).
\end{array}
\end{equation*}
Instead of \emph{adding} the positional embedding, we also explored concatenating it to the slot mix; we did not find large differences from doing so.

\subsection{ViT Encoders as Dense Features Extractors}
\label{app:vit}
The Vision Transformer (ViT)~\citep{Dosovitskiy2021ViT} architecture takes an input frame, denoted here as $\vx \in \mathbb{R}^{244 \times 224 \times 3}$, which is divided into a grid of non-overlapping contiguous patches of resolution $N \times N$. Each of these patches is then passed through a linear transformation to generate a set of patch feature embeddings, $\vh^{0} \in \mathbb{R}^{L \times D}$. 

The set of patch tokens is subsequently provided as input to a standard Transformer network. 
This Transformer network is comprised of a sequence of self-attention and feed-forward layers, alongside residual connections for each layer. 
These residual connections are important for letting each patch representation $\vh^{i}$ keep a correspondence to the original image patch representation $\vh^{0}$ and thus making the patch representation a dense (with the resolution $\sqrt{L}\times\sqrt{L}$) representation of the image.
In the self-attention layers, the token representations are updated through an attention mechanism that takes into account the representations of all tokens:
\begin{equation*}
\begin{array}{rlrl}
{[\mathbf{q}^{i}, \mathbf{k}^{i}, \mathbf{v}^{i}]} & =\mathbf{h}^{i-1} \mathbf{U}_{q k v} & \mathbf{U}_{q k v} & \in \mathbb{R}^{D \times 3 D} \\
\mA & =\operatorname{softmax}\left(\mathbf{q}^i \textbf{k}^{i \top} / \sqrt{D}\right) & \mA \in \mathbb{R}^{L \times L} \\
\mathbf{o}^{i} & =\mA \mathbf{v}^i
\end{array}    
\end{equation*}
Here $\mathbf{q}^{i}, \mathbf{k}^{i}, \mathbf{v}^{i}, \mathbf{o}^{i}$ are queries, keys, values and outputs of the self-attention layer $i$.
In contrast to DINOSAUR~\citep{seitzer2023bridging}, which uses the outputs $\mathbf{o}^{i}$ as the target image representation, we use attention keys $\mathbf{k}^{i}$ from the last self-attention layer of ViT as the dense image representation that is provided to the temporal similarity loss (see \fig{fig:keys} for a detailed comparison of these representations). 
However, we still use the outputs $\mathbf{o}^{i}$ as the input to the slot attention grouping module.

\subsection{Other Modules}
\label{app:our_model}
We group the dense encoder features with a recurrent slot attention module similar to \citet{Singh2022STEVE} and \citet{kipf2022conditional}. 
First, we transform the original features with a two-layer MLP with an output dimension equal to the slot dimension. 
Second, we use a slot attention module initialized with randomly sampled slots to group the first frame features, while for subsequent frames, we initialize the slot attention module with the slots of the previous frame, additionally transformed with a predictor module. 
We use the GRU recurrent unit in the slot attention grouping, but not the residual MLP.
Similar to SAVi~\citep{kipf2022conditional} and STEVE~\citep{Singh2022STEVE}, we use a one-layer transformer as the predictor module. 
In addition, we propose to decouple the number of Slot Attention iterations in the first frame and other frames of the video. 
This allows more iterations for the first frames (we use 3 iterations similar to image-based methods) and fewer iterations for the next frames where the initialization is much better (we use 2 iterations). 
For computational reasons, we were training on relatively short $4$-frame segments of original videos, \ie $T=4$.

\subsection{Next-Frame Feature Prediction Details}
\label{app:ffrec}
In this part, we cover implementation details for the next-frame feature prediction ablation presented in \tab{tab:loss_sim_alation_movic} and in \tab{tab:loss_sim_alation_ytvis}.
Reconstructing frame features from the current and next frame simultaneously with a single decoder is problematic because the decoder masks that are used for evaluation would be in reference to both the current and next frame. 
One way to overcome this problem is by using two decoder heads: $d_{\text{current}}$ for the current frame and $d_{\text{next}}$ for the next frame. 
Each head produces its own predictions and masks. 
In this case, masks from the $d_{\text{current}}$ head can be used for evaluation. 
While more powerful, this approach also requires more memory and is slower than standard setting with only one head. 
In our experiments, we confirm that the version with two different Mixer decoders performs better than simultaneous reconstruction with one decoder (see \tab{tab:ablation_decoders}).
We use this better version for our comparisons even though it is heavier than our method which needs only one decoder.

\begin{table}
\centering
\caption{Next-frame feature prediction with different decoders.}
\label{tab:ablation_decoders}
\begin{tabular}{@{}lcccccccc@{}} \toprule
& \multicolumn{2}{c}{MOVi-C} && \multicolumn{2}{c}{MOVi-E} && \multicolumn{2}{c}{YT-VIS} \\\cmidrule{2-3}\cmidrule{5-6}\cmidrule{8-9}
& FG-ARI & mBO && FG-ARI & mBO && FG-ARI & mBO \\
\midrule
One decoder head  & 44.6 & 23.5 && 61.3 & 22.1 && 33.4 & 24.6 \\
Two decoder heads & 47.2 & 24.7 && 62.9 & 24.0 && 37.9 & 27.3 \\
\bottomrule
\end{tabular}
\end{table}

\subsection{Baselines}
\label{app:baselines_details}

\paragraph{Block Pattern} 
The block pattern baseline serves to show metrics for a trivial decomposition of the video into regular blocks.
It is intended to show the difficulty of the dataset and how much object-centric methods improve upon such a trivial solution, as the metrics values could be difficult to interpret without further calibration.
To this end, we are splitting the video to $k$ spatial blocks consistently for all frames of the video, similar to how \citet{seitzer2023bridging} are splitting images into regular blocks. 

\paragraph{SAVi} We reimplement SAVi~\citep{kipf2022conditional} close to the official implementation\footnote{\url{https://github.com/google-research/slot-attention-video/}}. 
In particular, we use the {SAVi-L} architecture for all experiments.
This corresponds to the version using a ResNet-34 encoder as described by~\citet{kipf2022conditional}, and a CNN broadcast mixture decoder with 4 layers.
We apply the unconditional version of SAVi, using a fixed learned slot initialization instead.
We train the model for \numprint{200000} steps on all datasets, with a batch size of 64, using image reconstruction as the training signal.
For training, we use videos with 4 frames, with a single slot attention step per frame.

\paragraph{STEVE} We reimplement STEVE~\citep{Singh2022STEVE} close to the official implementation\footnote{\url{https://github.com/singhgautam/steve}}. 
We use the proposed configuration for the MOVi datasets and only change the number of slots to 11 for MOVi-C, 15 for MOVi-E and 7 for YT-VIS 2021.
STEVE trains a dVAE~\citep{rolfe2017discrete} on the video frames to extract a discrete latent code that is used as the reconstruction target.
STEVE uses a CNN encoder with 4 layers and a Transformer decoder with 8 layers.
We train the model for \numprint{200000} steps for MOVi-C and YT-VIS datasets and for  \numprint{100000} steps for MOVi-E datasets which resulted in optimal performance on this dataset, with a batch size of 24.
Like \citet{Singh2022STEVE}, for training, we use videos with 3 frames, with two slot attention steps per frame.

\subsection{Hyperparameters}
\label{app:hyperparams}
The full list of the hyperparameters used to train \method on MOVi-C, -D, -E, and YouTube-VIS 2021 datasets are presented in \Tab{tab:hyperparams}. Most of the hyperparameters are similar to previous methods, while new ones (such as softmax temperature $\alpha$ and time shift $k$) are optimized using a grid search.\looseness-1

\begin{table}[tbh!]
\caption{Hyperparameters of \method for the main results on MOVi-C, MOVi-E, and YouTube-VIS 2021 datasets.}
\label{tab:hyperparams}
\small
\centering
\setlength{\tabcolsep}{0.47em}
\begin{tabular}{@{}llrrr@{}}
\toprule
Dataset                            &                           & \textbf{MOVi-C}  & \textbf{MOVi-E}    & \textbf{YouTube-VIS} \\ \midrule
Training Steps                     &                           & 100k       & 100k       & 100k       \\
Batch Size                         &                           & 128         & 128         & 128        \\
Training Segment Length            &                           & 4        & 4         & 4       \\
LR Warmup Steps                    &                           & 2500      & 2500      & 2500      \\
Optimizer                           &                           & Adam    & Adam    & Adam     \\
Peak LR                            &                           & 0.0004     & 0.0004     & 0.0004     \\
Exp.~Decay               &                           & 100k       & 100k       & 100k       \\
ViT Architecture                   &                           & ViT Base      & ViT Base      & ViT Base      \\
Patch Size                         &                           & 8          & 8          & 16         \\
Feature Dim. $D_\text{feat}$       &                           & 768       & 768        & 768       \\
Gradient Norm Clipping             &                           & 0.05        & 0.05        & 0.05        \\ \midrule
Image/Crop Size                    &                           & 224        & 224        & 224        \\
Cropping Strategy                  &                           & Full       & Full       & Rand.~Cent.~Crop    \\
Augmentations                      &                           & --         & --         & Rand.~Hori.~Flip \\
Image Tokens                       &                           & 784      & 784       & 196  \\ \midrule
\multirow{4}{*}{Slot Attention}      &           Slots           & 11         & 15         & 7         \\
& Iterations (first / other frames)              & 3/2          & 3/2        & 3/2       \\
& Slot Dim.~$D_\text{slots}$       & 128        & 128        & 64        \\ \midrule
\multirow{3}{*}{Predictor}            & Type                      & Transformer        & Transformer      & Transformer \\
& Layers                    & 1         & 1         & 1          \\
& Heads                     & 4         & 4         & 4          \\ \midrule
\multirow{4}{*}{Decoder}            & Type                      & Mixer        & Mixer      & Mixer \\
& Allocator Transformer Layers                    & 2         & 2         & 3          \\
& Allocator Transformer Heads                     & 4         & 4         & 4          \\
& Renderer MLP Layers                    & 4         & 4         & 3          \\
& Renderer MLP Hidden Dim.               & 1024       & 1024       & 1024        \\    \midrule

\multirow{2}{*}{Loss}      & Softmax Temperature   $\tau$                & 0.075         & 0.075            & 0.25        \\
& Time-shift $k$               & 1        & 1        & 1          \\
& Feature reconstruction weight $\alpha$               & --        & --        & 0.1          \\
\end{tabular}
\end{table}

\subsection{Compute Requirements}
We used a cluster of A100 GPUs (with 40 and 80 Gb memory) for running the experiments. 
A single training run (100k steps) of \method{} equipped with DINO B/16 with a batch size of 128 takes roughly 18 hours on one A100 GPU with 40 GB memory. 
We use 5 seeds for the final results and 3 seeds for other experiments. 
Overall, we estimate the total compute spend on the whole project including training of all baselines and the method development (including dead ends) to be around 800-1000 GPU days (this is a rough estimate obtained from our cluster usage).

\section{Dataset Details}
In this section, we provide details about the datasets used in this work. 
See \tab{tab:dataset-overview} for an overview.

\begin{table*}
    \centering
    \caption{Overview of datasets used in this work.
    For the training process, we solely utilize images or videos derived from the relevant datasets, with no reliance on labels.
    To generate central crops, we initially resize the mask so that its shorter dimension is $224$ pixels. Subsequently, we extract the most centrally located crop, maintaining a size of $224$ by $224$ pixels.
    }
    \label{tab:dataset-overview}
    {\footnotesize
    \setlength{\tabcolsep}{0.28em}
    \begin{tabular}{@{}llllp{0.19\textwidth}@{}}
    \toprule
    Dataset & Videos &Images & Description & Citation \\
    \midrule
    MOVi-C, -D, -E & \numprint{9750} & -- & Train split videos & \citet{Greff2021Kubric} \\
    MOVi-C, -D, -E validation & \numprint{250}& -- & Val.~split w.~instance segm.~labels & \citet{Greff2021Kubric} \\
    COCO 2017 & -- &\numprint{118287} & Train split & \citet{Lin2014COCO} \\
    COCO 2017 validation & -- &\numprint{5000} & Val split w.~instance segm.~labels & \citet{Lin2014COCO} \\
    YouTube-VIS 2021 & 2785 & -- & Part of train split videos & \citet{vis2021} \\
    YouTube-VIS 2021 & 210 & -- & Part of train split w.~instance segm.~labels & \citet{vis2021} \\
    YouTube-VIS 2019 & 200& -- & Part of train split w.~instance segm.~labels  & \citet{vis2019} \\
    DAVIS-2017-Unsupervised & 30& -- & Val. split  w.~instance segm.~labels  & \citet{davis2017} \\
    \bottomrule
    \end{tabular}}
\end{table*}

\paragraph{MOVi datasets}
For MOVi-C, MOVi-D and MOVi-E, we use the standard training and test splits provided in the respective releases of MOVi datasets: $9750$ training videos and $250$ validation videos. 
For a fair comparison, all the methods are trained on images with the same initial resolution $128 \times 128$. 
Consistent with previous work~\citep{kipf2022conditional, elsayed2022savi++, Singh2022STEVE, seitzer2023bridging}, we utilize the validation split of the MOVi dataset for our evaluations.

\paragraph{YouTube-VIS datasets}
The YouTube-VIS datasets~\citep{vis2019, vis2021} are benchmarks originally used for \emph{supervised} video instance segmentation. 
The video instance segmentation task involves segmenting, tracking and classifying instances throughout a video. 
It is a challenging dataset because it contains various different classes of objects and the complexities of real-world video dynamics. 
To the best of our knowledge, this is the first work attempting \emph{unsupervised} instance segmentation on YouTube-VIS.

There are two different versions of this dataset: YouTube-VIS 2019 and YouTube-VIS 2021. 
The YouTube-VIS 2019 dataset is mainly derived from the Video Object Segmentation (VOS) dataset. 
It has a limited number of individual instances (an average of $1.7$ per video for the training set), and the categories of instances in the same video are usually different. 
In contrast, the 2021 edition of YouTube-VIS incorporates a higher quantity of objects with more difficult trajectories (average $3.4$ per video for the additional videos in the train set) and thus are more interesting for object-centric learning. We sample both training and validation frames with a rate of 6 frames per second (each 5th frame of the original videos with $30$ \emph{fps}).

As the original validation dataset is not publicly available, and the evaluation server does not compute the object-centric metrics we need, we split the original training set into two parts ($210$ videos for validation and the other videos for training).
Overall, we use $2775$ videos for training, and $210$ videos additionally added in YT-VIS 2021 train for validation.  

\paragraph{DAVIS dataset} The DAVIS-2017-Unsupervised dataset~\citep{davis2017} is used for video object segmentation and also contains videos with multiple objects per video (average is $1.97$ for 30 validation videos). 
As those videos are not related to YouTube-VIS videos, this dataset is useful for the evaluation of transfer abilities for real-world object-centric learning algorithms. 

\paragraph{COCO dataset} To test the properties of our proposed similarity loss for image-based object-centric learning we use the COCO dataset~\citep{Lin2014COCO}. Similar to \citet{seitzer2023bridging}, we use \emph{118287} training images (without labels) to train and \numprint{5000} validation images (with labels) for performance evaluation.

\section{Additional Examples}

We include additional example predictions of our model:

\begin{itemize}
    \item \Fig{fig:ytvis_examples}: comparing \method to STEVE on Youtube-VIS 2021.
    \item \Fig{fig:movi_c_examples}: comparing \method to STEVE on MOVi-C.
    \item \Fig{fig:movi_e_examples}: comparing \method to STEVE on MOVi-E.
    \item \Fig{fig:loss_diff}: comparing feature reconstruction and temporal similarity loss on MOVi-C.
    \item \Fig{fig:ytvis2021_examples_long}: long-term video predictions on Youtube-VIS 2021 (\method with DINO B/16 features).
    \item \Fig{fig:ytvis2021_examples_long_dinov2}: long-term video predictions on Youtube-VIS 2021 (\method with DINOv2 B/14 features).
\end{itemize}

\setcounter{figure}{0}

\label{app:visualizations}
\begin{figure}
  \centering
  \includegraphics[width=\textwidth]{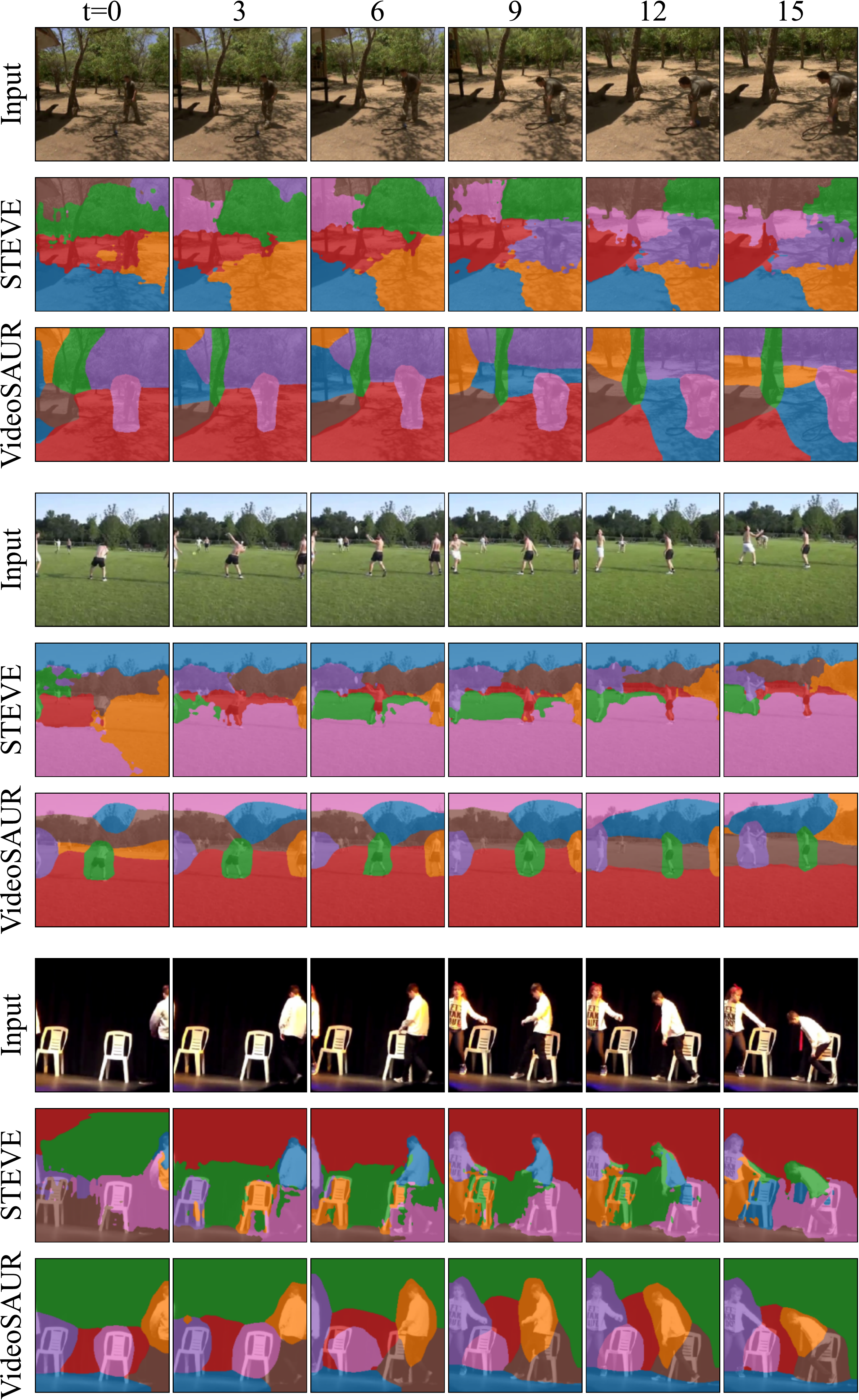}
    \caption{Additional examples on YouTube-VIS 2021.}
    \label{fig:ytvis_examples}
\end{figure}

\begin{figure}
  \centering
  \includegraphics[width=\textwidth]{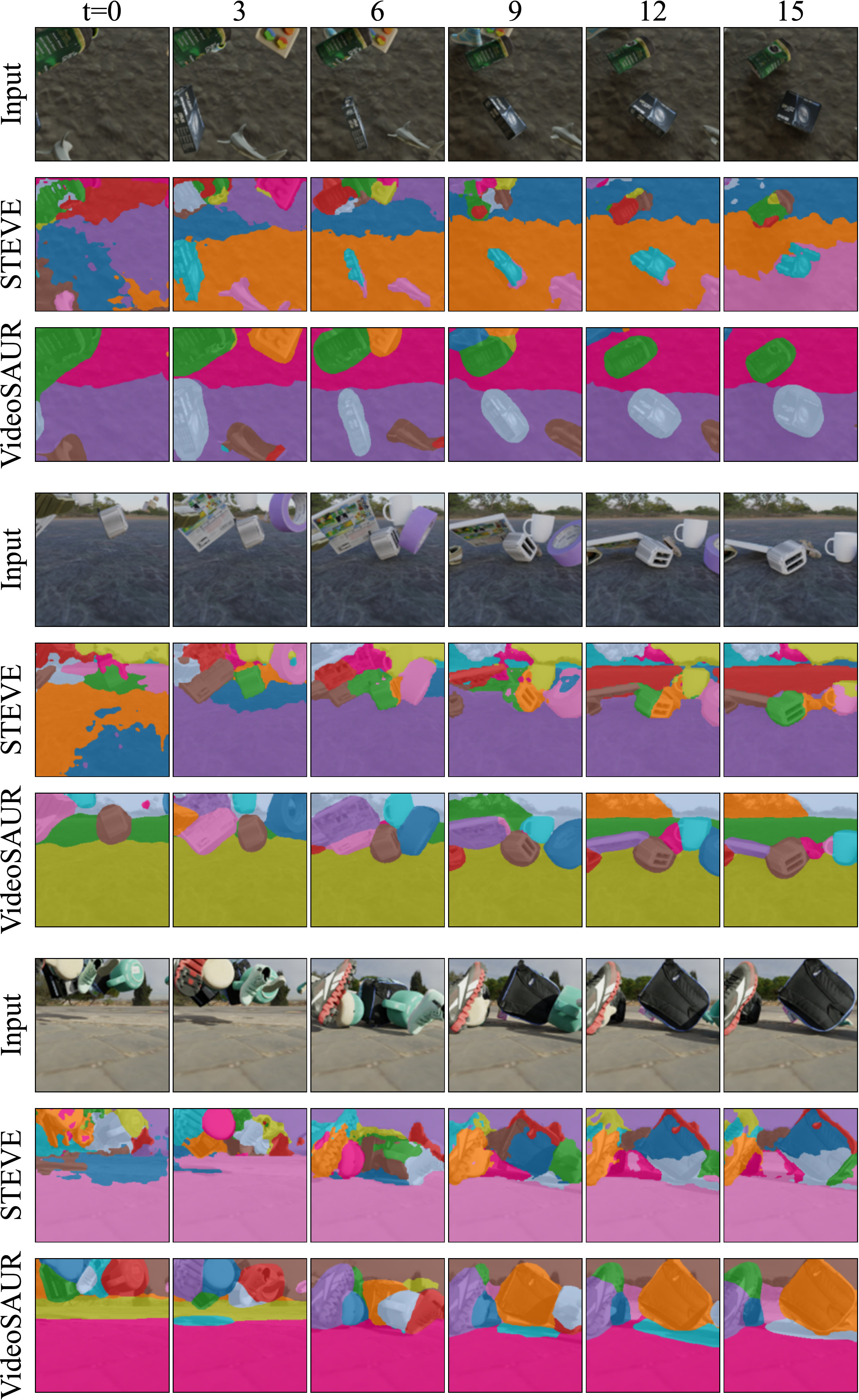}
    \caption{Additional examples on MOVi-C.}
    \label{fig:movi_c_examples}
\end{figure}

\begin{figure}
  \centering
  \includegraphics[width=\textwidth]{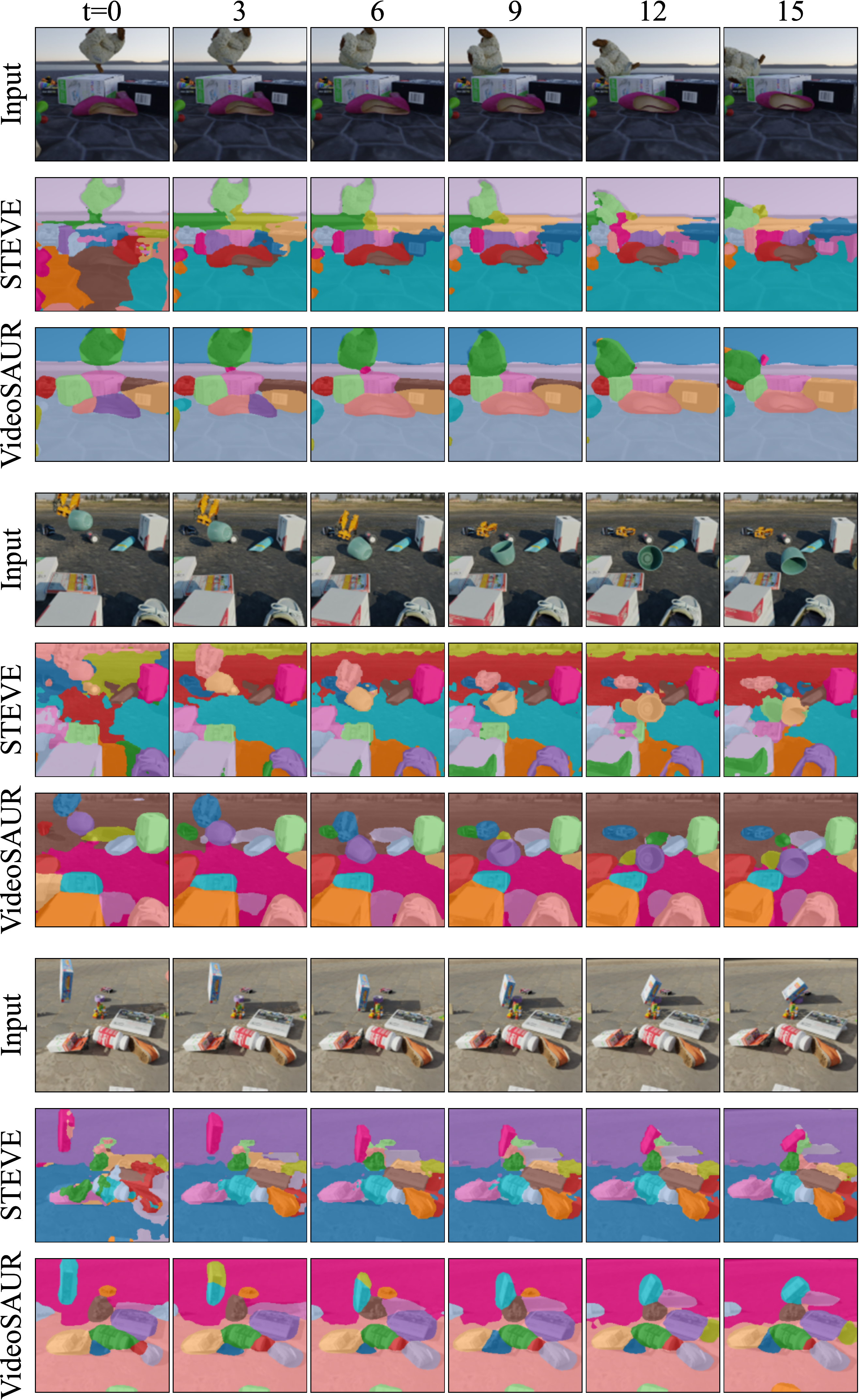}
    \caption{Additional examples on MOVi-E.}
    \label{fig:movi_e_examples}
\end{figure}

\newpage
\begin{figure}[t]
  \centering
  \includegraphics[height=0.95\textheight]{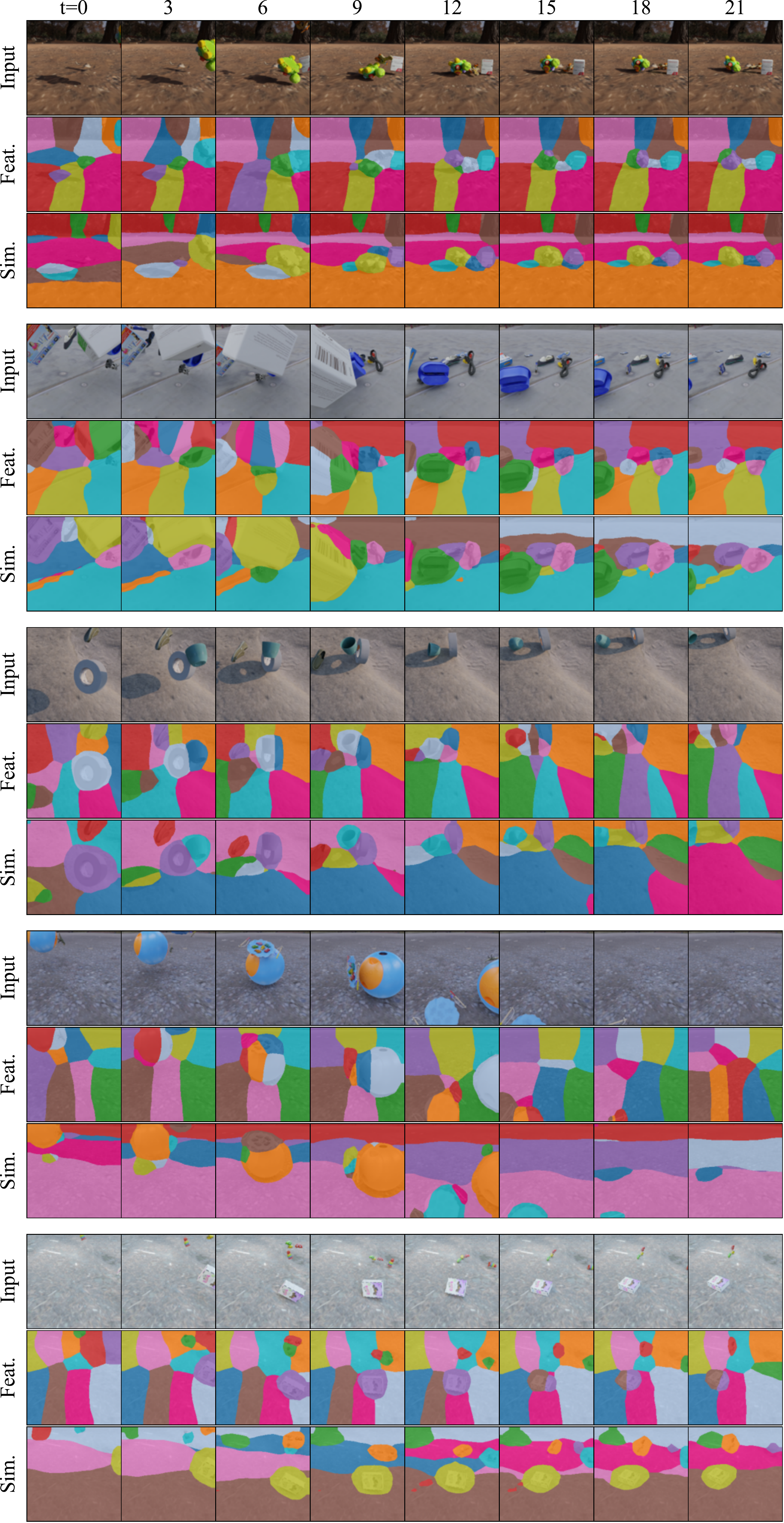}
    \caption{Difference between \method train with feature reconstruction and temporal feature similarity losses on MOVi-C. We show videos with larger differences in performance between the two methods.}
    \label{fig:loss_diff}
\end{figure}

\begin{figure}[t]
  \centering
  \includegraphics[height=\textheight]{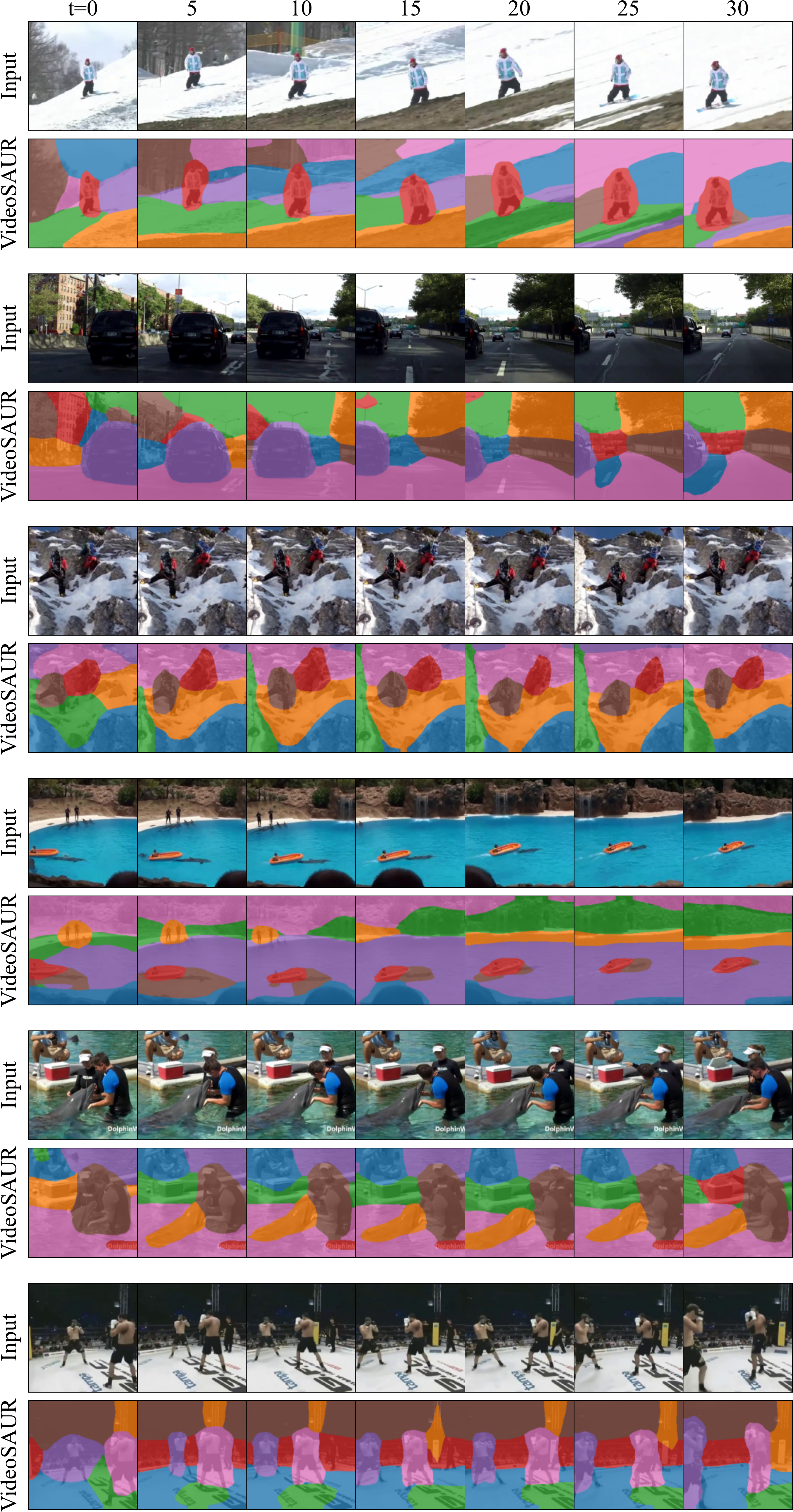}
    \caption{Prediction of \method with DINO B/16 features on longer videos from YouTube-VIS 2021.}
    \label{fig:ytvis2021_examples_long}
\end{figure}

\begin{figure}[t]
  \centering
  \includegraphics[height=\textheight]{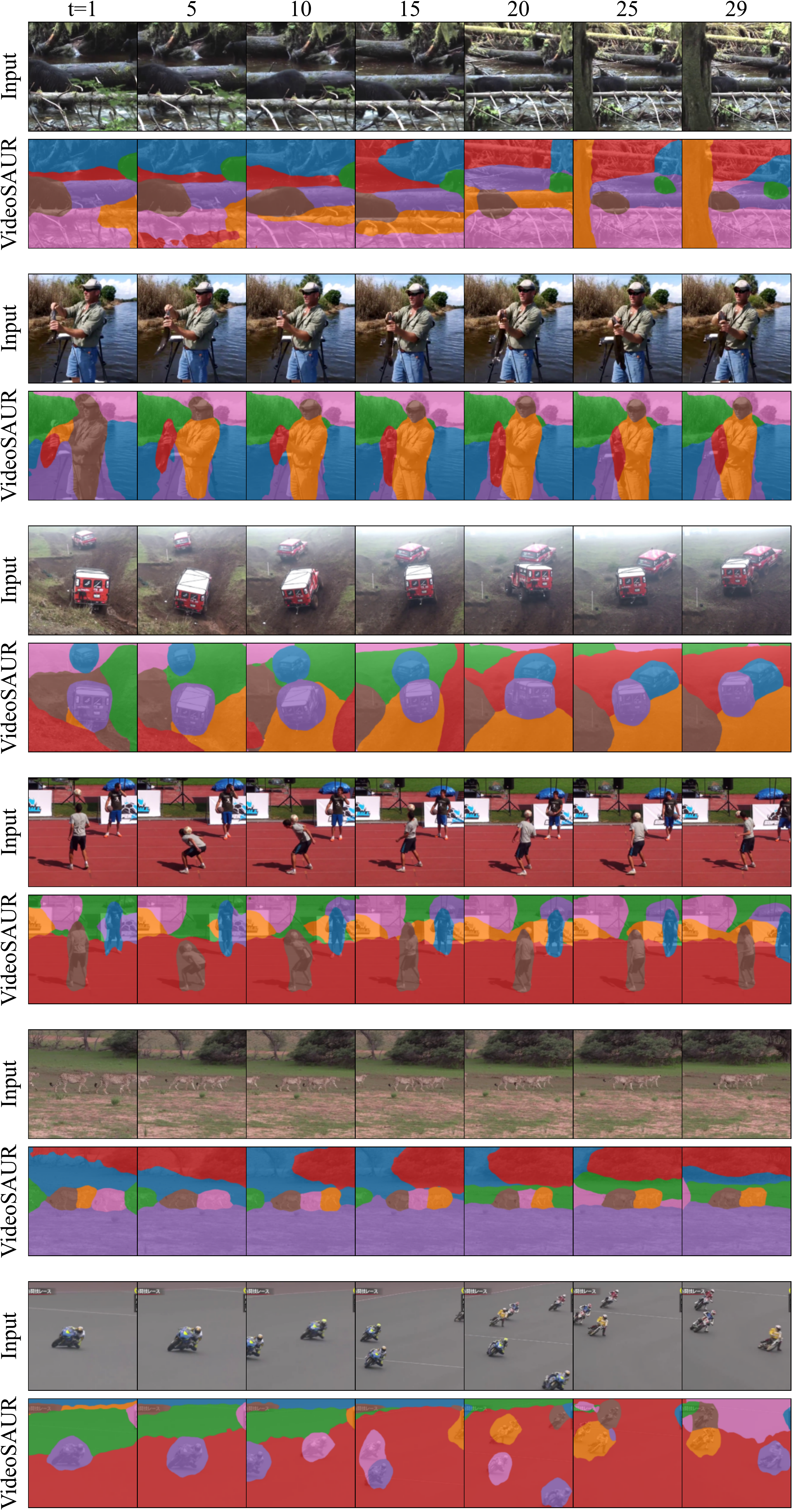}
    \caption{Prediction of \method with DINOv2 B/14 features on longer videos from YouTube-VIS 2021.}
    \label{fig:ytvis2021_examples_long_dinov2}
\end{figure}

\end{document}